\documentclass{article}

\usepackage{microtype}
\usepackage{graphicx}
\usepackage{subcaption}
\usepackage{booktabs} 

\usepackage{hyperref}

\usepackage{tabularx}
\usepackage[most]{tcolorbox}
\usepackage{xcolor}

\definecolor{sympycolor}{HTML}{66c2a5}
\definecolor{nlcolor}{HTML}{8da0cb}
\definecolor{smtcolor}{HTML}{b3b3b3}

\definecolor{tealheading}{HTML}{66c2a5}
\definecolor{tealbody}{HTML}{e6f5f0}  
\newtcolorbox{promptboxSYM}[1]{%
  enhanced,
  breakable,
  colback=tealbody,
  colframe=tealheading,
  boxrule=0.6pt,
  arc=2mm,
  left=4mm,right=4mm,top=2mm,bottom=2mm,
  fonttitle=\bfseries,
  title={Prompt #1},
  before skip=8pt,
  after skip=10pt
}

\definecolor{coralheading}{HTML}{b3b3b3}
\definecolor{coralbody}{HTML}{e0e0e0}
\newtcolorbox{promptboxSMT}[1]{%
  enhanced,
  breakable,
  colback=coralbody,
  colframe=coralheading,
  boxrule=0.6pt,
  arc=2mm,
  left=4mm,right=4mm,top=2mm,bottom=2mm,
  fonttitle=\bfseries,
  title={Prompt #1},
  before skip=8pt,
  after skip=10pt
}

\definecolor{blueheading}{HTML}{8da0cb}
\definecolor{bluebody}{HTML}{e8ecf5}  

\newtcolorbox{promptboxNL}[1]{%
  enhanced,
  breakable,
  colback=bluebody,
  colframe=blueheading,
  boxrule=0.6pt,
  arc=2mm,
  left=4mm,right=4mm,top=2mm,bottom=2mm,
  fonttitle=\bfseries,
  title={Prompt #1},
  before skip=8pt,
  after skip=10pt
}

\usepackage{listings}
\usepackage{caption} 

\usepackage[preprint]{icml2026}

\usepackage{amsmath}
\usepackage{amssymb}
\usepackage{mathtools}
\usepackage{amsthm}

\usepackage{xcolor}
\definecolor{Preserve}{RGB}{55, 94, 151}      
\definecolor{Implicit}{RGB}{171, 73, 64}      
\definecolor{Injected}{RGB}{64, 127, 101}     
\definecolor{Ordering}{RGB}{118, 83, 148}     

\newcommand{\pos}[1]{\textcolor{green!60!black}{#1}}

\usepackage{adjustbox}

\usepackage{colortbl}

\definecolor{lightgray}{gray}{0.92}
\usepackage{etoc}
\usepackage{minitoc}

\usepackage{titletoc}

\usepackage{xspace}
\newcommand{\ourmethod}{\texttt{Opt-Sym}\xspace}

\usepackage{adjustbox}
\usepackage{multirow}
\usepackage{enumitem}

\usepackage[capitalize,noabbrev]{cleveref}

\theoremstyle{plain}

\theoremstyle{definition}

\theoremstyle{remark}

\usepackage[textsize=tiny]{todonotes}

\icmltitlerunning{Adaptive Problem Generation via Symbolic Representations}

\begin{document}

\twocolumn[
  \icmltitle{Adaptive Problem Generation via Symbolic Representations}


  \icmlsetsymbol{equal}{*}

  \begin{icmlauthorlist}
    \icmlauthor{Teresa Yeo}{smart}
    \icmlauthor{Myeongho Jeon}{kaist}
    \icmlauthor{Dulaj Weerakoon}{smart}
    \icmlauthor{Rui Qiao}{smart}\\
    \icmlauthor{Alok Prakash}{smart}
    \icmlauthor{Armando Solar-Lezama}{mit}
    \icmlauthor{Archan Misra}{smu}
  \end{icmlauthorlist}

  \icmlaffiliation{kaist}{Korea Advanced Institute of Science and Technology}
  \icmlaffiliation{smart}{Singapore-MIT Alliance for Research and Technology}
  \icmlaffiliation{mit}{Massachusetts Institute of Technology}
  \icmlaffiliation{smu}{Singapore Management University}

  \icmlcorrespondingauthor{Teresa Yeo}{teresa.yeo@smart.mit.edu}

  \icmlkeywords{Machine Learning, ICML}

  \vskip 0.3in
]



\printAffiliationsAndNotice{}  

\begin{abstract}
We present a method for generating training data for reinforcement learning with verifiable rewards to improve small open-weights language models on mathematical tasks. 
Existing data generation approaches rely on open-loop pipelines and fixed modifications that do not adapt to the model’s capabilities. Furthermore, they typically operate directly on word problems, limiting control over problem structure.
To address this, we perform modifications in a symbolic problem space, representing each problem as a set of symbolic variables and constraints (e.g., via algebraic frameworks such as SymPy or SMT formulations). 
This representation enables precise control over problem structure, automatic generation of ground-truth solutions, and decouples mathematical reasoning from linguistic realization. We also show that this results in more diverse generations.
To adapt the problem difficulty to the model, we introduce a closed-loop framework that learns modification strategies through prompt optimization in symbolic space. Experimental results demonstrate that both adaptive problem generation and symbolic representation modifications contribute to improving the model’s math solving ability.
\end{abstract}

\section{Introduction}

Recent advances in large language models (LLMs)~\cite{singh2025openaigpt5card,hubert2025alphaproof} have demonstrated significant progress on mathematical tasks. Unfortunately, there is a large and growing gap between the capabilities of closed frontier models and those of smaller open-weights models. This is a problem for many applications that are either cost sensitive or where privacy concerns prevent the use of closed models. Improving the mathematical 
abilities of these smaller models is therefore an important goal. 

Recent work has shown that it is possible to scale mathematical reasoning through synthetic training data for reinforcement learning from verifier feedback (RLVR). Effective training requires both verifiable solutions for reward computation~\cite{guo2025deepseek,lambert2024tulu} and problem distributions that matches the model's capabilities~\cite{zeng2025rlve,razin2023vanishing,razin2025makes}.
Unfortunately, most existing methods generate problems through open-loop pipelines that cannot easily adapt to a student model's evolving capabilities or to different student models~\cite{yu2023metamath,tang2024mathscale,toshniwal2024openmathinstruct1,toshniwal2024openmathinstruct2,tong2024dart}. These approaches can generate samples that are either too easy or too hard, reducing training efficiency.

\begin{figure*}[t]
    \centering
    \includegraphics[width=0.9\textwidth]{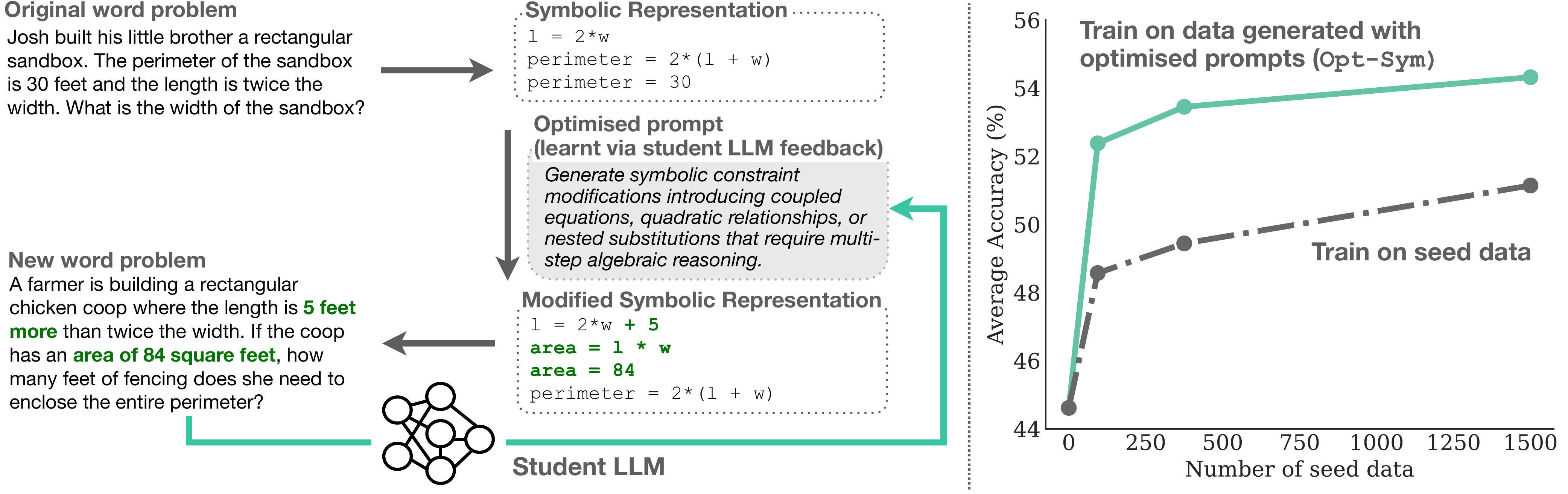}
    \caption{\textit{\textbf{Left:}} \textbf{Symbolic Data Generation with Prompt Optimization.} We generate training data for mathematical tasks by modifying problems in symbolic space rather than directly in natural language. Starting from an original word problem (top left), we convert it to a symbolic representation (top right). An optimized prompt (middle, italics)—learned through closed-loop feedback from the model's performance—specifies how to modify the symbolic structure to generate \textit{appropriately} challenging problems. We call these prompts \ourmethod. The modified symbolic representation (bottom right) is then translated back to a new word problem (bottom left). The LLM responses guide prompt optimization (green loop) to produce problems that are challenging. This symbolic approach enables precise control over problem structure and automatic ground-truth generation. 
    In addition to introducing coupled equations, nested substitutions, note how the new word problem also returns a different story line than the original, i.e., generations through symbolic representations returns in more diverse training data. See~\cref{sec:add-analysis} for a more detailed analysis.
    \textbf{\textit{Right:}} \textbf{Training on data generated via \ourmethod improves model performance.} We show the performance from training on data from \ourmethod (green) compared to training on the original seed data (grey). The y-axis shows the performance averaged over several benchmarks, and the x-axis shows the amount of seed data used. \ourmethod achieves an 8\% improvement in performance even with as little as 100 seed data points. This is compared to the baseline's 4\% improvement. This demonstrates both the effectiveness and data efficiency of our approach.
    See~\cref{tab:results_main,tab:results_average_3b} and~\cref{fig:results-ablations} for the full results and analysis.
    }
    \label{fig:intro}
\end{figure*}

We propose a framework for generating training data by modifying symbolic representations of mathematical expressions in a closed-loop manner (see \cref{fig:intro}). Operating on a symbolic representation decouples mathematical semantics from linguistic variations, enabling precise control over problem complexity (operation types, equation depth) while independently varying lexical presentation.
%
Specifically, we 
\textbf{1.} parse mathematical problems into symbolic representations (\cref{sec:method-representations}), 
\textbf{2.} apply learned prompts to these symbolic representations to generate modifications (\cref{sec:data-gen}), and 
\textbf{3.} use the student model's output as feedback to adaptively learn and refine these prompts (\cref{sec:closed-loop}). 
Unlike open-loop generation, this closed-loop process provides fine-grained control over problem difficulty and semantic properties and  ensures that the generated problems remain \textit{appropriately challenging}.

Our work makes two primary contributions. First, we introduce a closed-loop data generation framework that adapts symbolic representations to generate new problems. This offers several advantages, e.g., automatic generation of ground-truth solutions and decoupling of the mathematical structure from the linguistic realization (\cref{sec:method}). Second, experiments on multiple benchmarks and models demonstrate that our approach consistently improves performance when trained with RLVR using either PPO or GRPO, compared to training with the seed data alone (\cref{sec:experiments}).




\section{Related Work}

\textbf{Symbolic representations for generating math problems.}
Several recent approaches leveraged symbolic representations to generate synthetic mathematical problems at scale. AlphaGeometry~\citep{trinh2024solving} generated 100 million synthetic theorems by sampling random diagrams and exhaustively deriving relationships using a symbolic deduction engine.~\citet{li2024neuro} formalized problems in SMT-LIB format and applied Markov Chain Monte Carlo sampling to generate 620K verified examples. MathCAMPS~\citep{mishra2025next} encoded mathematical skills as formal grammars from which symbolic problems are sampled and then realized as natural language. GSM-Symbolic~\cite{mirzadeh2024gsm} introduced a benchmark by converting GSM8K problems into templates with perturbable variations. However, these methods rely on fixed generation procedures—hand-designed mutation operators, static grammars, or fixed prompts—that are not adapted to the model's capabilities. We build on this by introducing prompt optimization over symbolic representations in a closed-loop framework.

\textbf{Open-loop Data Generation} creates data without feedback from the model. There are several approaches. First, question bootstrapping and rephrasing approaches create diverse problems through systematic augmentation strategies. MetaMath~\cite{yu2023metamath} introduced forward-backward augmented reasoning and multi-perspective question rewriting to generate the MetaMathQA dataset, while MathScale~\cite{tang2024mathscale} constructed concept graphs from seed problems and generates new questions via random walks over mathematical concepts. MuggleMath~\cite{li2024mugglemath} and MuMath~\cite{you2024mumath} systematically studied query evolution strategies including numerical variation, complexity enhancement, and multi-perspective reformulation. Second, rejection sampling and verification methods focus on solution quality through automated filtering and learning. DART-Math~\cite{tong2024dart} addressed sampling bias through difficulty-aware allocation that samples proportionally to problem difficulty. Math-Shepherd~\cite{wang2024math} developed automatic process reward models for step-level verification, and recent work demonstrates that even incorrect solutions can improve learning when processed with per-step credit assignment~\cite{setlur2024rl}. Key-Point-Driven Data Synthesis~\cite{huang2025key} generated problems from mathematical concepts through structured prompt frameworks, showing that conceptual scaffolding improves problem diversity and coverage. Third, large-scale synthesis approaches combine these principles with hybrid reasoning formats to create massive training datasets. MAmmoTH~\cite{yue2023mammoth} pioneered chain-of-thought and program-of-thought integration across diverse mathematical domains, while ToRA~\cite{gou2023tora} and MathCoder~\cite{wang2023mathcoder} generated interactive tool-use trajectories interleaving natural language with code execution. OpenMathInstruct-1~\cite{toshniwal2024openmathinstruct1} and OpenMathInstruct-2~\cite{toshniwal2024openmathinstruct2} scaled to 1.8M and 14M problem-solution pairs respectively using open-weight models, with Skywork-Math~\cite{zeng2024skywork} investigating data scaling laws up to 2.5M instances. JiuZhang3.0~\cite{zeng2024skywork} demonstrated efficient synthesis by distilling GPT-4's generation capability into smaller 7B models. 
In contrast to these works, our approach \textit{learns} to generate targeted training data for a given student model.

%
\textbf{Closed-loop Data Generation with Student Model Feedback.}
Recent work has explored adaptive data generation where generators modify their strategies based on the model's performance. STaR~\cite{zelikman2022star} introduced bootstrapping methods where models iteratively generate solutions, filter by correctness, and fine-tune on successful attempts. LLM2LLM~\cite{lee2024llm2llm} targeted student model weaknesses by generating synthetic examples similar to incorrectly answered problems, showing substantial improvements in low-data regimes.
DataEnvGym~\cite{khan2024dataenvgym} framed this as sequential decision-making, where a teacher agent creates training data based on student model errors or weak skills, improving students across multiple domains. RLVE~\cite{zeng2025rlve} used adaptive verifiable environments that procedurally generate problems and dynamically adjust difficulty to match evolving policy capabilities, demonstrating that environment scaling across 400 manually engineered environments improves reasoning performance.
Montessori-Instruct~\cite{li2024montessori} used influence functions to identify high-impact training data, training the teacher with DPO to optimize for student-specific learning preferences. Our work differs by combining closed-loop optimization with symbolic problem representations, enabling automated solution generation via prompt-based modifications 
See~\cref{sec:closed-loop} for more details.

\textbf{Prompt Optimization Methods.} Early methods like APE~\cite{zhou2022large} and OPRO~\cite{yang2023large} framed prompt discovery as search problems where LLMs generate candidate prompts evaluated against task objectives. PromptAgent~\cite{wang2023promptagent} extended this through strategic planning with Monte Carlo Tree Search, discovering expert-level prompts by iteratively refining based on error feedback. TextGrad~\cite{yuksekgonul2024textgrad} implemented automatic differentiation via text, backpropagating natural language feedback to optimize prompts in compound AI systems. While these methods are generally used to improve task performance, our work adapts this framework, in particular Textgrad, to generate useful training data, using the model's output as the optimization signal. See~\cref{sec:closed-loop} for details.

\section{Method}\label{sec:method}

\subsection{Problem Setup}
We address the problem of generating targeted training data for improving the math capabilities a given student model.

\textbf{Notations.} Let $\mathcal{D}_{\text{seed}} = {(q_i, a_i)}_{i=1}^{N}$ denote a seed dataset of mathematical problems, where $q_i$ is a problem statement and $a_i$ is its corresponding solution. Our framework employs an {LLM}, $M_{\text{gen}}$ to generate new problem–solution pairs. 
We use the prompts $\mathcal{P}^{\text{Sym}} \triangleq \{ p_{\text{opt}}^{\text{Sym}}, p_{\text{render}} \}$ to generate data, where $p_{\text{opt}}^{\text{Sym}}$ generates the modified symbolic problem and $p_{\text{render}}$ converts a symbolic problem into natural language. Note that the answer to this modified problem is attained by calling a solver e.g., SymPy or Z3 in the case of SMT-LIB representations.
Our main baseline, modifying the word problem directly uses the prompts $\mathcal{P}^{\text{NL}} \triangleq \{ p_{\text{opt}}^{\text{NL}}, p_{\text{solve}} \}$, where $p_{\text{opt}}^{\text{NL}}$ generates the modified word problem and $p_{\text{solve}}$ prompts $M_{\text{gen}}$ to solve the modifed word problem.
Both $p_{\text{opt}}^\text{Sym}$ and $p_{\text{opt}}^\text{NL}$ are learnable prompts that controls how seed problems are modified.
The resulting synthetic problem-solution pairs are later used to train a student LLM {policy model $\pi_{\text{student}}$} with RLVR.

\textbf{Objective.} Our goal is to learn a prompt $p_{\text{opt}}^{\text{Sym}}$ that produces targeted training data for a given student model. 
We achieve this via prompt optimization, where the prompts are adapted based on the current skill level of the student model. This creates a closed feedback loop that targets the weaknesses of the model.

\subsection{Representations for Math Problems}\label{sec:method-representations}

A key design choice in our framework is the use of symbolic representations as an intermediate form for problem generation, i.e., we perform closed-loop adaptation in this symbolic space, rather than on the word problems directly.
This provides several benefits:

\begin{itemize}[label={}, leftmargin=1em, topsep=0pt, partopsep=0pt, itemsep=0.5pt, parsep=0pt]
\item \textit{Controllability and structure.} Symbolic representations make the mathematical structure explicit and controllable, allowing precise specification of generation constraints in prompts (e.g., ``include fractions in coefficients'', ``ensure multi-step solutions''). In our experiments, optimized prompts for symbolic generation ($p_{\text{opt}}^\text{Sym}$) evolve to include such structural constraints, enabling systematic control and exploration of the problem space that is difficult to achieve through natural language prompts alone.
\item \textit{Automatic solution verification.} Symbolic solvers enable immediate verification that generated problems are well-posed and solvable, and provide corresponding solutions automatically. 
\item \textit{Decoupling math from language.} By separating symbolic problem generation from natural language rendering, we decouple mathematical content creation from linguistic presentation, allowing targeted optimization of $p_{\text{opt}}^\text{Sym}$ while keeping $p_{\text{render}}$ fixed.
{This decoupling enables the generation of mathematically and lexically diverse problems.}
\end{itemize}

\textbf{Symbolic representations.}
We primarily adopt \textit{SymPy representations} to encode the mathematical structure of problems. SymPy provides a flexible symbolic language for representing mathematical expressions and equations, supporting exact arithmetic as well as symbolic transformation of both expressions and solutions. 
We also consider \textit{SMT-LIB representations}, which enable constraint-based problem generation; however, we find that rendering SMT problems into natural language often results in unnatural-sounding word problems. 
Accordingly, we use SymPy as our main symbolic representation, and show results with SMT-LIB as an additional analysis in~\cref{sec:add-analysis}.

\subsection{Data Generation Pipeline}
\label{sec:data-gen}

\begin{figure*}[t]
    \centering
    \includegraphics[width=0.95\textwidth]{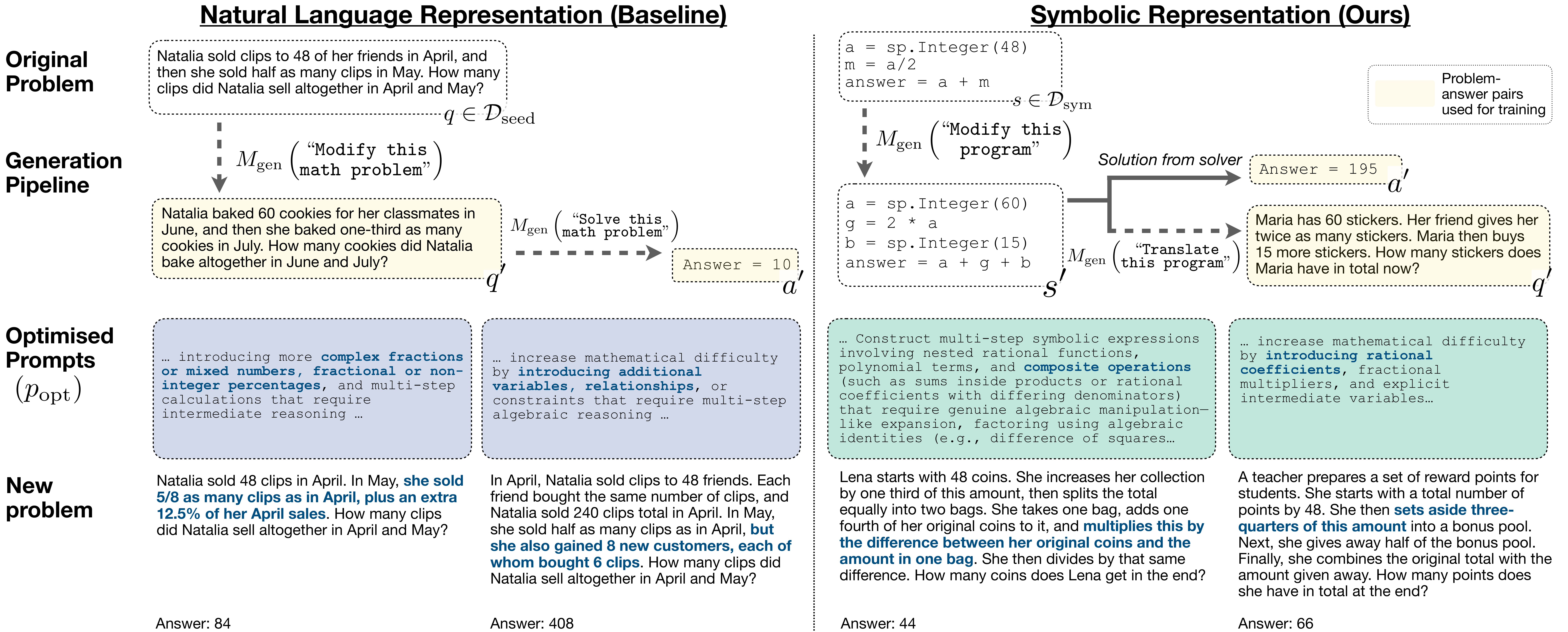}
    \caption{\textbf{\textit{Top:}} 
    \textbf{Data Generation Pipeline.} We generate new math problems from seed examples using two approaches. 
    \textit{Left (Natural Language (NL) representation):} Given a seed problem $q \in \mathcal{D}_{\text{seed}}$, we prompt LLM $M_{\text{gen}}$ to create a modified problem $q'$, then prompt it again to generate its solution. 
    \textit{Right (Symbolic Representation):} Given a seed problem $s \in \mathcal{D}_{\text{sym}}$, prompt $M_{\text{gen}}$ to modify this representation to create $s'$, translate $s'$ back to natural language to obtain $q'$, and use a solver to compute the answer. Both pipelines produce problem-answer pairs for training.\\
    \textit{\textbf{Bottom:}} \textbf{Examples of optimized prompts and generated problems}. 
    Each column shows a different optimized prompt $p_{\text{opt}}$ (top, truncated) applied to the same seed problem shown at the top. The resulting generated problem is show below. The blue text highlights the modifications requested by each prompt. These prompts are learned through closed-loop optimization (see~\cref{sec:closed-loop}) that takes student model performance as feedback, enabling them to target specific weaknesses. See~\cref{sec:app-full-opt-prompts} for complete optimized prompts across different settings. 
    The optimized prompts for both natural language and symbolic representations suggest similar mathematical modifications (fractions, multi-step reasoning, additional variables), indicating some overlap in how problem difficulty should be increased for the student model. However, the actual generated problems exhibit distinct characteristics. The NL modifications produce problems with complex narratives that require careful interpretation, while symbolic modifications yield problems with explicit mathematical operations stated directly in the text. Thus, the space in which modifications are applied impacts the final problem structure and complexity.
    }
    \label{fig:method-data-gen-pipeline}
    \vspace{-1em}
\end{figure*}

We now describe our data generation pipeline, which produces training data via symbolic representations that generates new math problems (\cref{fig:method-data-gen-pipeline}-right). We also describe the pipeline for our baseline, that generates problems directly in natural language (\cref{fig:method-data-gen-pipeline}-left). Both pipelines are guided by prompts that are optimized through our closed-loop framework (see~\cref{sec:closed-loop}).



\textbf{Symbolic generation pipeline.} The symbolic pipeline operates as follows:
\begin{itemize}[label={}, leftmargin=1em, topsep=0pt, partopsep=0pt, itemsep=0.5pt, parsep=0pt]
\item \textbf{0.} \textit{Seed data preparation.} For the symbolic pipeline, we first translate problems from $\mathcal{D}_{\text{seed}}$ (e.g., GSM8K, MATH) into a symbolic representation of each seed problem to get $\mathcal{D}_{\text{sym}}$. This translation is \textbf{performed once} as a preprocessing step. The natural language pipeline uses the original seed problems directly (see~\cref{fig:method-data-gen-pipeline}). 
\item \textbf{1.} \textit{Symbolic problem generation.} Given a symbolic problem from $\mathcal{D}_{\text{sym}}$, we apply $M_{\text{gen}}$ with prompt $p_{\text{opt}}^\text{Sym}$ to generate a new problem. 
To get the ground-truth answer, we solve the symbolic problem using the corresponding solver i.e., Sympy or Z3 for the case of SMT-LIB.
\item \textbf{2.} \textit{Natural language rendering.} The generated symbolic problem is then converted to a natural language word problem using $M_{\text{gen}}$ with $p_{\text{render}}$. This prompt instructs the model to create a coherent, natural-sounding problem statement that corresponds to the symbolic representation. 
\end{itemize}

\textbf{Natural language generation pipeline.} For comparison, we also implement a natural language pipeline that generates problems directly:

\begin{itemize}[label={}, leftmargin=1em, topsep=0pt, partopsep=0pt, itemsep=0.5pt, parsep=0pt]
\item \textbf{1.} \textit{Direct problem generation.} Given a seed problem $q_i \in \mathcal{D}_{\text{seed}}$, we use $M_{\text{gen}}$ with prompt $p_{\text{opt}}^{\text{NL}}$ to directly generate a new word problem in natural language.
\item \textbf{2.} \textit{Solution generation.} We use $M_{\text{gen}}$ with prompt $p_{\text{solve}}$ to generate a solution.
\end{itemize}


\subsection{Closed-Loop Data Generation via Prompt Optimization}
\label{sec:closed-loop}

A key component of our framework is the closed-loop optimization of generation prompts based on student model performance. Rather than using hand-crafted prompts for generating data, we \textit{learn} the prompt to create problems.

\textbf{Optimization framework.} We optimize prompts using TextGrad~\citep{yuksekgonul2024textgrad}, a ``textual'' gradient-based optimization method for natural language. For the symbolic pipeline, we optimize a single prompt $p_{\text{opt}}^\text{Sym}$ that modifies SymPy code to create new problems {in the symbolic representation space}. For the natural language pipeline, we optimize a single prompt $p_{\text{opt}}^{\text{NL}}$ that modifies natural language problems directly. The optimization process proceeds iteratively: \textbf{1.} generate a batch of problems using the current prompt, \textbf{2.} evaluate the student model on these problems, \textbf{3.} compute a text-based loss signal, \textbf{4.} use TextGrad to update the prompt. 
This allows the prompt to learn to generate problems that targets the student's weaknesses.

\textbf{Loss function.} The loss function guides the prompt optimization by evaluating the quality of generated problems. For the symbolic pipeline, given a SymPy problem $s_i$ from $\mathcal{D}_{\text{sym}}$, a generated SymPy problem $s'_j$ produced using the current prompt, the student model's output $o_j$ on the rendered natural language version of $s'_j$, and the correct answer $a'_j$ obtained from the SymPy solver, we define a text-based loss:
\begin{equation}
\begin{split}
&\mathcal{L}_{\text{sym}}\bigl(s_i, s'_j, o_j, a'_j\bigr)\\
&\overset{\text{def}}{=}\mathcal{E}_{\text{LLM}}\!\left(
s_i,\; s'_j,\; o_j,\; a'_j
\;\middle|\;
\mathcal{C}_{\text{diff}},\; \mathcal{C}_{\text{guard}}
\right),
\end{split}
\end{equation}
{where $\mathcal{E}_{\text{LLM}}$ denotes an LLM-based evaluator that produces a text-based loss signal by assessing whether the generated problem meets a target difficulty criterion $\mathcal{C}_{\text{diff}}$ and satisfies a set of quality guardrails $\mathcal{C}_{\text{guard}}$.}
For the natural language pipeline, the loss takes the original natural language problem $q_i$, generated problem $q'_j$, student output $o_j$, and correct answer $a'_j$ (by calling $M_\text{gen}$ with $p_\text{solve}$):
\begin{equation}
\begin{aligned}
&\mathcal{L}_{\text{NL}}\bigl(q_i, q'_j, o_j, a'_j\bigr)\\
&\overset{\text{def}}{=}
\mathcal{E}_{\text{LLM}}\!\left(
q_i,\; q'_j,\; o_j,\; a'_j
\right. \left.
\middle|\;
\mathcal{C}_{\text{diff}},\; \mathcal{C}_{\text{guard}}
\right).
\end{aligned}
\end{equation}
The loss consists of a difficulty criterion and guardrails:

\textit{Difficulty.} The generated problem should be challenging enough that the student model produces an incorrect answer. Problems that are too easy (student answers correctly) provide less training signal, while appropriately difficult problems help the student learn from mistakes.

\textit{Guardrails.} The loss also enforces several constraints to ensure problem quality and prevent degenerate solutions. 
These include: appropriate length (similar to seed problems), 
avoiding unrelated contexts) and proper structure (clean encoding without redundant operations). 
These guardrails prevent the optimization from generating trivially difficult but low-quality problems, such as extremely long or poorly formatted problems that are hard for the wrong reasons. See~\cref{app:prompt-opt-details} for the exact loss used.


\textbf{Optimization dynamics.} The prompt, $p_{\text{opt}}$ evolves over multiple iterations.
Initially, $p_{\text{opt}}$ may produce problems similar to the seed data. The optimization progresses as follows, 
\begin{align}
g^{(t)} &= \mathrm{ComputeTextGradient}\!\left(p_\text{opt}^{(t)}, \mathcal{L}^{(t)}, \mathcal{D}_s\right) \\
p_\text{opt}^{(t+1)} &= \mathrm{UpdatePrompt}\!\left(p_\text{opt}^{(t)}, g^{(t)}\right)
\end{align}
where $\mathcal{D}_s\in\{\mathcal{D}_\text{Sym},\mathcal{D}_\text{Seed}\}$. After several steps, $p_{\text{opt}}$ learns the modifications that target the weaknesses of the student model. For the symbolic pipeline, $p_{\text{opt}}$ naturally returns detailed specifications (e.g., ``include fractions", ``ensure multi-step solutions") as discussed in Section 3.2. The optimization process discovers these modifications automatically and does not require manual engineering.


\textbf{Stopping criteria.} We terminate the optimization when: \textbf{1.} the maximum number of iterations is reached, or \textbf{2.} the generated problems achieve a target difficulty level (e.g., student accuracy drops below a threshold).

\subsection{Data Validation and Student Model Training.}

After generating problems through the optimized prompts, we apply validation filters to ensure data quality before using the problems for training. We then train the student model on the filtered dataset using reinforcement learning.

\textbf{Validation and filtering.} The generated problems are passed to an LLM, e.g., GPT-5-mini~\cite{singh2025openaigpt5card} to check if the answers are correct. Both pipelines produce similar rates of valid problems. Problems that fail these checks are replaced by a problem generated via a simple prompt e.g., ``Modify this problem" on the same seed data. 

\textbf{Student model training.} We train the student policy model $\pi_{\text{student}}$ using reinforcement learning from verifiable rewards (RLVR). Specifically, we use either Proximal Policy Optimization (PPO)~\cite{schulman2017proximal} or Group Relative Policy Optimization (GRPO)~\cite{shao2024deepseekmath} with a binary accuracy reward: 1 for producing the correct answer and 0 otherwise. The training dataset consists of the filtered generated problems combined with the original seed data from $\mathcal{D}_{\text{seed}}$. This combination ensures that the student model learns from both the challenging generated problems and maintains coverage of the seed distribution.

\begin{table*}[!t]
\centering
\caption{\textbf{Performance comparison across different generated data types.} Accuracies~(\%) on several benchmarks before (Data type None) and after training the 1.5B student model with different combinations of generated data types and RL methods. Seed-data (+MATH) represents training using both GSM8K and MATH datasets, while \ourmethod (+MATH) further augments this setting by adding optimized problems derived from both GSM8K and MATH. 
\ourmethod outperforms the baselines on averaged performance with either PPO or GRPO.
Furthermore, \ourmethod results in consistent improvement in performance across benchmarks compared to training with seed data only ($\Delta$ (vs. Seed) rows).
For the results on the 3B model, see \cref{tab:results_average_3b} (averaged) and~\cref{tab:results_main_3b} (full).}
\begin{adjustbox}{max width=\linewidth}
\begin{tabular}{ll>{\columncolor{lightgray}}c lllllll}
\toprule
\textbf{RL Method} & \textbf{Data type} & \textbf{Average} & \textbf{GSM8K} & \textbf{GSM-Sym} & \textbf{Sym-p1} & \textbf{Sym-p2} & \textbf{GSM-Plus} & \textbf{MATH-500} & \textbf{AIME24} \\
\midrule

 & None
 & 44.61 & 69.75 & 64.16 & 49.98 & 23.48 & 50.67 & 50.91 & 3.33 \\
\midrule

\multirow{6}{*}{\textbf{PPO}}
 & Seed-data   & 51.14 & 79.61 & 72.96 & 61.12 & 32.16 & 54.25 & 54.53 & 3.33 \\
 & Base-NL     & 53.03 & 80.52 & 76.22 & 64.10 & 35.68 & 58.83 & 52.52 & 3.33 \\
 & Base-Sym    & 53.14 & 80.74 & 74.74 & 64.38 & \textbf{36.64} & 59.71 & \textbf{55.73} & 0.00 \\
 & Opt-NL      & 53.16 & 80.89 & 76.04 & 62.64 & 34.88 & \textbf{60.58} & 53.72 & 3.33 \\
 & \ourmethod  & \textbf{54.32} & \textbf{81.58} & \textbf{76.70} & \textbf{64.94} & 35.36 & 59.83 & 55.13 & \textbf{6.67} \\
\cmidrule(lr){2-10}
 & $\Delta$ (vs. Seed)
 & \pos{+3.18} & \pos{+1.97} & \pos{+3.74} & \pos{+3.82}
 & \pos{+3.20} & \pos{+5.58} & \pos{+0.60} & \pos{+3.34} \\
\midrule

\multirow{11}{*}{\textbf{GRPO}}
 & Seed-data  & 51.40 & 79.45 & 73.04 & 61.54 & 32.64 & 55.08 & 54.73 & 3.33 \\
 & Base-NL    & 53.27 & 80.97 & 75.08 & 63.70 & 32.48 & 59.04 & 54.93 & 6.67 \\
 & Base-Sym   & 52.58 & 79.76 & 75.56 & 63.60 & 33.96 & 59.63 & 55.53 & 0.00 \\
 & Opt-NL     & 52.60 & 79.53 & 74.22 & 63.88 & 34.48 & 60.38 & \textbf{55.73} & 0.00 \\
 & \ourmethod & \textbf{54.71} & \textbf{81.50} & \textbf{77.52} & \textbf{65.58} & \textbf{35.40} & \textbf{61.38} & 54.93 & \textbf{6.67} \\
\cmidrule(lr){2-10}
 & $\Delta$ (vs. Seed)
 & \pos{+3.31} & \pos{+2.05} & \pos{+4.48} & \pos{+4.04}
 & \pos{+2.76} & \pos{+6.30} & \pos{+0.20} & \pos{+3.34} \\
\cmidrule(lr){2-10}
 & Seed-data (+MATH)  & 52.37	 & 79.45	 & 72.54 & 60.40	 & 30.24	 & 57.33	 & 59.96	 & 6.67 \\
 & Base-Sym (+MATH)   & 54.33	 & 81.12	 & 77.36	 & 63.92	 & 33.96	 & 60.33	 & 56.94	 & 6.67 \\
 & \ourmethod (+MATH) & \textbf{56.18}	 & \textbf{81.96}	 & \textbf{78.22}	 & \textbf{65.78}	 & \textbf{38.00}	 & \textbf{61.50}	 & \textbf{61.17}	 & \textbf{6.67} \\
 
\cmidrule(lr){2-10}
 & $\Delta$ (vs. Seed + MATH)
 & \pos{+3.81}	& \pos{+2.5}	& \pos{+5.68}	& \pos{+5.38}	& \pos{+7.76}	& \pos{+4.17}	& \pos{+1.21}	& \pos{+0} \\
\bottomrule
\end{tabular}
\end{adjustbox}
\label{tab:results_main} \vspace{-1em}
\end{table*}

\section{Experiments}
\label{sec:experiments}

We now evaluate our approach on several mathematical benchmarks. Our code is available at \url{https://github.com/aserety/adaptive-problem-generation}.

\subsection{Experimental Setup}
We use Qwen2.5-1.5B-Instruct and Qwen2.5-3B-Instruct~\cite{qwen2025qwen25technicalreport} as our student models and GSM8K~\citep{cobbe2021training} and MATH~\citep{hendrycks2measuring} as seed datasets.

\textbf{Prompt Optimization.} We optimize prompts using Textgrad. The prompt is initialized to “Modify this problem/program”. For the 1.5B model, we run for a total of 4 steps, and 10 steps for the 3B model, or until the stopping criterion is reached. The batch size is 16. We use GPT-5-mini as the optimizer in Textgrad. See~\cref{sec:app-full-opt-prompts} for the full optimized prompts for the different student models.

\textbf{Data Generation.} Unless otherwise specified, we generate data from 1500 datapoints sampled from the GSM8K seed dataset. For Opt-NL and \ourmethod, we optimize 4 prompts and generate one datapoint per prompt per seed example, yielding $1500\times4$ generated datapoints in total. To introduce variation across these optimized prompts, we apply different restrictions to the prompt optimization loss e.g., beyond the difficulty criterion and guardrails, we constrain modifications to be local changes. The exact loss formulations are detailed in~\cref{app:prompt-opt-details}. The seed-data baseline trains exclusively on these same seed datapoints. Note that all methods incorporate the seed data in their training sets. We set $M_{\text{gen}}=\text{GPT-5-mini}$.

\textbf{RLVR Training.} We train student models using reinforcement learning from verifiable rewards with the VERL framework~\citep{sheng2025hybridflow}. The reward signal is based solely on solution accuracy. We train the models with two algorithms for comparison: Group Relative Policy Optimization (GRPO)~\cite{shao2024deepseekmath} and Proximal Policy Optimization (PPO)~\cite{schulman2017proximal} and report the results from both. {Implementation details are shown in Section~\ref{app:rlvr-details}.}

\textbf{Evaluation.} We evaluate the trained models on the test set of GSM8K, variants of GSM-Symbolic~\cite{mirzadeh2024gsm}, GSM-Plus~\cite{li2024gsm}, MATH500~\cite{hendrycks2measuring,lightman2023letsverifystepstep}, 
and AIME24~\citep{he2024olympiadbench}. The results are attained with greedy decoding (temperature=0). We report the \textbf{pass@1 accuracy}. 

\textbf{Baselines.} We compare against the follow baselines. The student model trained with: 
\begin{itemize}[label={}, leftmargin=1em, topsep=0pt, partopsep=0pt, itemsep=0.5pt, parsep=0pt]
\item \textit{Seed-data:} Seed dataset $\mathcal{D}_{\text{seed}}$ only. Establishes performance without data augmentation.
\item \textit{Base-NL:} Data generated by modifying the word problem directly with the prompt ``Modify this word problem." Tests whether simple prompting for NL-level augmentation improves performance.
\item \textit{Base-Sym:} Data generated by modifying the symbolic problem with the prompt ``Modify this program." Tests if simple prompting at the symbolic level is effective.
\item \textit{Opt-NL:} Data generated with the NL optimized prompts $p_{\text{opt}}^{\text{NL}}$. Isolates the benefit of prompt optimization without symbolic manipulation.
\end{itemize}

\textbf{Our Approach: \ourmethod.} Our method trains the student model using data generated with optimized prompt $p_{\text{opt}}^{\text{Sym}}$.

\subsection{Performance Across Generation Methods}

{\cref{tab:results_main} shows the performance of the 1.5B student model trained with data generated by different methods. Across both RL algorithms, \ourmethod consistently achieves the highest average accuracy and delivers stronger gains than natural-language modifications (Opt-NL) or their non-optimized variants (Base-NL, Base-Sym), demonstrating the effectiveness of combining symbolic representations with optimized prompts. When additional MATH seed data is included, \ourmethod (+MATH) continues to improve performance on GSM8K, GSM-Sym, and Sym-p1/p2, indicating robustness across different seed datasets, despite drops on MATH-500 and GSM-Plus. 

\cref{tab:results_average_3b} shows the averaged results for the 3B model (see~\cref{tab:results_main} for the full breakdown). \ourmethod outperforms the baselines with GRPO, and achieves competitive performance with PPO.


\begin{table}[h]
\centering
\caption{\textbf{Averaged accuracy (\%) for the 3B model.} \ourmethod outperforms the baselines with GRPO and achieves competitive performance with PPO. See~\cref{tab:results_main_3b} for a breakdown on the different benchmarks.}
\begin{adjustbox}{max width=\linewidth}
\begin{tabular}{lllllll}
\toprule
\textbf{Average} & None & Seed-data & Base-NL & Base-Sym & Opt-NL & \ourmethod \\ 
\midrule
\textbf{PPO}  & \multirow{2}{*}{59.35} & 60.47 & 62.22 & 61.07 & \textbf{63.10} & 62.50 \\
\textbf{GRPO} &  & 62.00 & 63.36 & 62.02 & 62.57 & \textbf{64.39} \\
\bottomrule
\end{tabular}
\end{adjustbox}
\label{tab:results_average_3b}
\vspace{-1.5em}
\end{table}

    \begin{figure*}[h]
    \centering
    \includegraphics[width=0.95\textwidth]{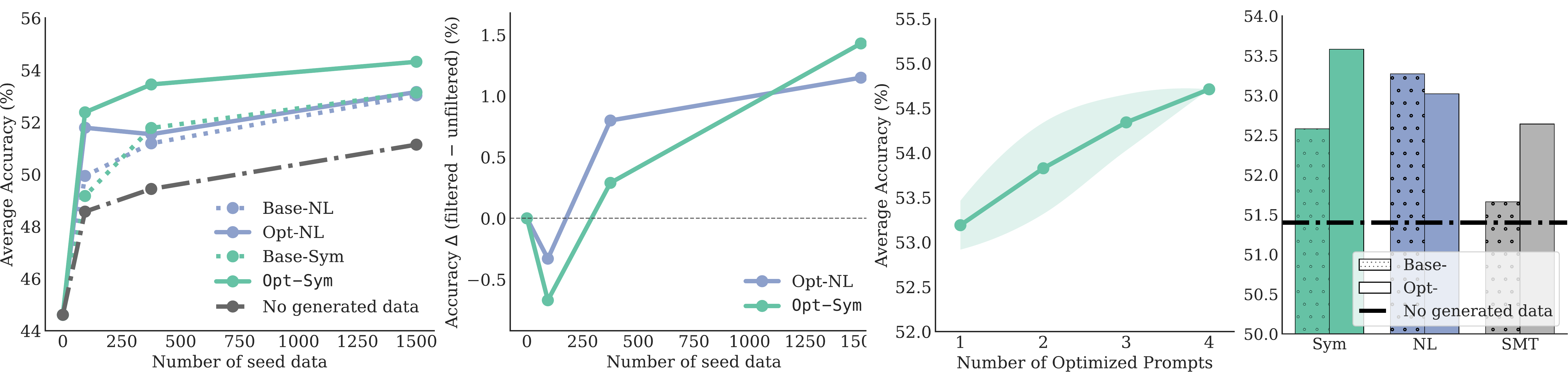}
        \caption{\textbf{Additional analysis.} 
        \textbf{a.} \textit{Data generation methods across subset sizes}: The 1.5B model trained with \ourmethod (green) outperforms the baseline approaches when trained on different data subsets. 
        \textbf{b.} \textit{Data filtering impact}: Filtering the generated training data improves performance on larger dataset sizes. This can be because for small datasets, filtering has a larger impact on data diversity, which can harm performance~\cite{setlur2024rl}.
        \textbf{c.} \textit{Prompt scaling}: Performance increases monotonically with the number of prompts used during data generation. This indicates that having more diverse data helps improve performance (see~\cref{fig:results-data-analysis} for further analysis on data diversity). As we optimize for 4 prompts in total, there are $4\choose n$ options for selecting each of the $n$ prompts etc. This shows the averaged accuracy (\%) and standard deviation over the different combinations. 
        \textbf{d.} \textit{Representation sensitivity}: The choice of mathematical problem representation significantly affects model performance. 
        }
        \label{fig:nb}
    \label{fig:results-ablations} 
\end{figure*}

\subsection{Additional Analysis}\label{sec:add-analysis}

\textbf{Generating data with optimized prompts results in higher data efficiency.} 
\cref{fig:nb} shows the performance where the student model was trained (with PPO) on data generated from different subsets of seed data. Opt-NL and \ourmethod achieve higher performance than Base-NL and Base-Sym in the low data regime, demonstrating that prompt optimization improves sample efficiency. See~\cref{fig:full_subset_results} for the full results with other RL methods, with and without data filtering and for the 3B student model.

\textbf{Performance increases with the number of optimized prompts used for generation.} 
We train on data generated using varying numbers of optimized prompts for \ourmethod. As we optimize for a total of 4 prompts, there are $4 \choose 1$ ways of selecting 1 prompt, $4 \choose 2$ ways of selecting 2 prompts etc. \cref{fig:results-ablations} shows the average and standard deviation over these combinations. Performance increases with the number of optimized prompts used, indicating that greater prompt diversity leads to more effective data augmentation.

\textbf{Representation matters: SymPy vs SMT-LIB vs NL.} 
We compare the performance of training the student model on different symbolic representations in~\cref{fig:nb}. The data generated via SMT-LIB results in lower performance compared to both SymPy and NL representations. A possible reason, discussed in \cref{sec:method-representations}, is that SMT-LIB's syntax translates word problems that tend to sound more unnatural and do not improve the student model's performance. 
\cref{app:representation-comparison} shows some examples. 

SMT solvers require explicit encoding of low-level computational details such as type conversions and integer arithmetic operations, which translate into verbose procedural descriptions in natural language. In contrast, SymPy allows mathematical relationships to be expressed at a higher level of abstraction, resulting in word problems that focus on relationships rather than implementation details. This difference makes SymPy generated problems sound more natural, while SMT-generated problems often expose computational mechanics that are irrelevant to problem-solving, thus, generating data that is less useful for training.


\textbf{Effect of filtering the generated data.}
We analyze the effect of filtering invalid or incorrect generated problems in~\cref{fig:nb}. We use GPT-5-mini to determine if the generated problem has the correct answer. If it does not, we replace that datapoint with its corresponding baseline data, i.e., a datapoint from Opt-NL gets replaced with one with the same seed from Base-NL.

Data filtering hurts performance when the student model is trained with small subsets of data ($\sim100$ seed datapoints). When the dataset is small, filtering reduces training data diversity, and it has been shown that the student model can benefit more from exposure to varied examples (even with some errors) than from a limited set of verified correct ones~\cite{setlur2024rl}. On the full generated dataset, filtering improves performance by 1-1.5\%.

\begin{table}[H]
\vspace{-1mm}
\centering
\small
\caption{\textbf{Averaged accuracy (\%) across multiple iterations of generation and fine-tuning on the 1.5B model.} \ourmethod shows consistent improvement as iterations increase. Base-Sym results with the same number of prompts per iteration are shown for comparison.}
\begin{tabular}{lcccc} 
\toprule
\textbf{Data Type} & \textbf{Iter. 0} & \textbf{Iter. 1} & \textbf{Iter. 2} & \textbf{Iter. 3} \\
\midrule
Base-Sym & \multirow{2}{*}{51.40} & 52.54 & 52.58 & 54.30 \\
\ourmethod &  & 53.90 & 54.71 & 56.26 \\
\bottomrule
\end{tabular}
\label{tab:multi-iter}
\vspace{-3mm}
\end{table}
\textbf{Multiple iterations of prompt optimization, data generation and fine-tuning.}
Our previous results employed only a single iteration of prompt optimization, data generation and fine-tuning. Here, we investigate if multiple iterations yield further performance gains. We performed this experiment with the 1.5B model and optimized for 2 prompts each iteration, accumulating training data from previous iterations at each step.~\cref{tab:multi-iter} shows the results for 3 iterations. \ourmethod achieves steady improvement, with accuracy increasing from 53.90\% to 56.26\%.

\subsection{Analysis of Generated Data}

\textbf{Generated problems are more diverse.} 
To measure the diversity of our generated data, we selected 100 seed problems and generated 10 variants of each of them. We then compute the embeddings of these problems with the sentence embedding model, all-MiniLM-L6-v2 model\footnote{\small\url{https://huggingface.co/sentence-transformers/all-MiniLM-L6-v2}}. For each seed problem, 
we compute the pairwise cosine distance for the generated variations. This is then averaged over all seed problems.~\cref{fig:results-data-analysis}~(left) shows this result for \ourmethod and Opt-NL. We also show this measure for the baseline prompts, Base-Sym and Base-Opt for comparison. Data generated via \ourmethod has an average pairwise cosine distance of 0.37 versus Opt-NL's 0.07. Furthermore, even using a simple baseline prompt, Base-Sym results in higher average cosine distance i.e., more diverse data than Base-NL. 

We show examples of the generated variants in~\cref{app:generated-problem-variations}. Opt-NL, which operates directly on word problems, produces variants that closely mimic the seed problem's surface features, i.e., it tends to retain the same characters, objects, and narrative context while making modifications to the mathematical operations. For example, all Opt-NL variants of a problem about ``Liza buying butter for cookies" maintain Liza, butter, and the cookie types, differing in details like the order of operations or minor additions.
In contrast, \ourmethod converts problems to SymPy representations, abstracting away these superficial features. This enables \ourmethod to generate variants with completely different protagonists, objects, and contexts while also making modifications to the mathematical structure, resulting in greater diversity both semantically and mathematically. See  ~\cref{app:qualitative-analysis} for more fine-grained qualitative analysis.

\begin{figure}[H] \vspace{-0.2em}
    \centering 
    \includegraphics[width=0.48\textwidth]{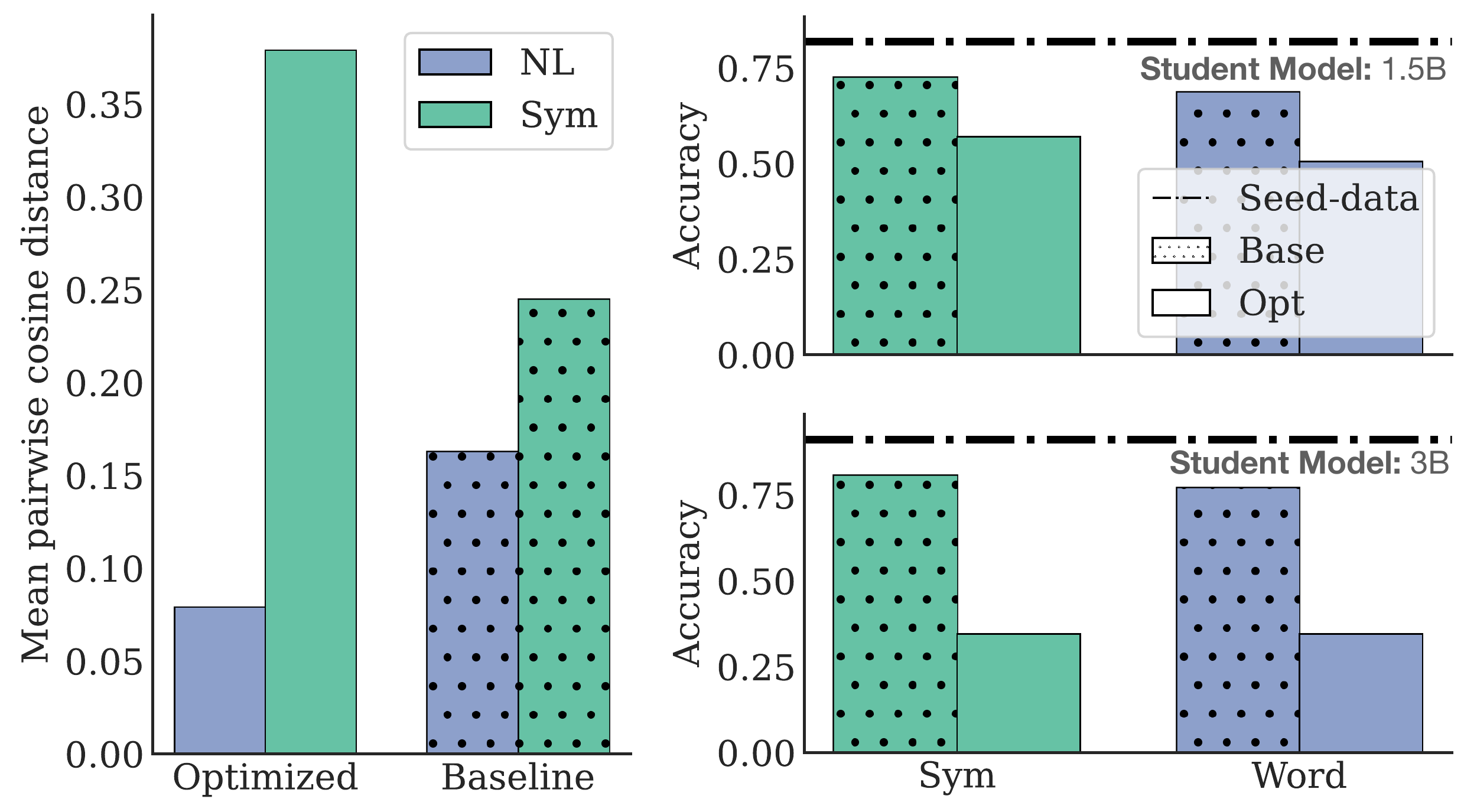}
    \caption{\textbf{Analysis of generated data}. \textbf{\textit{Left:}} Generation via symbolic representations produce more diverse data. For each seed problem, we generate 10 variants using baseline and optimized prompts ($\mathcal{P}^{\text{NL}}$ and $\mathcal{P}^{\text{Sym}}$), compute their embeddings, and measure average pairwise cosine distance. Optimized prompts with symbolic representations (Sym) achieve substantially higher diversity than natural language (NL) variants.
    \textit{\textbf{Right:}} Student models (1.5B and 3B) achieve lower accuracy on generated data compared to the seed data (dash-dot line), with optimized prompts producing the most challenging problems. Accuracy is similar across word-based and symbolic generation methods. The difficulty gaps from seed data suggests successful problem variations.
    } %
    \vspace{-1em}
    \label{fig:results-data-analysis}
\end{figure}

\textbf{Difficulty of the generated problems.} 
\cref{fig:results-data-analysis} shows the accuracy of the student models on the (filtered) generated problems \textit{before training}. \ourmethod generates problems that are more challenging compared to seed and baseline generated data, indicating \textit{effective augmentations} to the dataset that are \textit{tailored to the student model}.

\section{Conclusion and Future Work}

We introduced a closed-loop framework for generating training data that adapts to student model performance through prompt optimization via symbolic representations. 
Empirical results show that our approach produces challenging and diverse problems that are useful for training student language models.
These are some of the limitations:
\vspace{-2pt}
\begin{itemize}[label={}, leftmargin=1em, topsep=0pt, partopsep=0pt, itemsep=0.5pt, parsep=0pt]
\item \textit{Symbolic Representation Dependency.} Our approach relies on converting problems into symbolic representation. This limits applicability to problems that allows for reliable formalization, i.e., problems involving ambiguous language, implicit constraints, or complex real-world reasoning may be difficult to capture symbolically.
\item \textit{Feedback Quality of Student Model.} The prompt optimization relies on student LLM output as the training signal, if the student model has idiosyncratic error patterns, optimization may overfit to these patterns rather than genuinely improving problem difficulty.
\item \textit{Reliance on proprietary LLMs.} These models have concerns over reproducibility, cost, etc. Furthermore, as we use GPT-5-mini for generation, filtering etc., the performance of the student model may be bounded by the capabilities of this model.
\item \textit{Extensions to other domains.} Although the proposed framework is conceptually applicable to other  domains with formal representations (e.g., logic problems via Prolog or planning tasks using PDDL), our evaluation is restricted to mathematical settings.
\end{itemize}

\textbf{Acknowledgments.} This research was supported by the National Research Foundation (NRF), Prime Minister’s Office, Singapore under its Campus for Research Excellence and Technological Enterprise (CREATE) programme. Myeongho Jeon was supported by the InnoCORE Program of the Ministry of Science and ICT (No. N10250156).

\textbf{Impact Statement.} This paper presents work whose goal is to advance the field of machine learning. There are many potential societal consequences of our work, none of which we feel must be specifically highlighted here.


\bibliography{example_paper}
\bibliographystyle{icml2026}

\newpage
\appendix
\onecolumn

\section*{Appendix Contents}
\vspace{0.8em}

{
\noindent
\textbf{A. \hyperref[sec:appendix:detail]{Implementation Details}}
\dotfill \pageref{sec:appendix:detail}

\vspace{0.2em}
\hspace{2.0em}
\hyperref[app:rlvr-details]{A.1 \ RLVR}
\dotfill \pageref{app:rlvr-details} 

\hspace{2.0em}
\hyperref[app:prompt-opt-details]{A.2 \ Prompt Optimization}
\dotfill \pageref{app:prompt-opt-details}

\hspace{2.0em}
\hyperref[app:data-generation]{A.3 \ Data Generation}
\dotfill \pageref{app:data-generation}

\vspace{0.2em}
\rule{\linewidth}{0.2pt}
\vspace{0.2em}

\noindent
\textbf{B. \hyperref[app:qualitative-analysis]{Qualitative Analysis}}
\dotfill \pageref{app:qualitative-analysis}

\vspace{0.2em}
\rule{\linewidth}{0.2pt}
\vspace{0.2em}

\noindent
\textbf{C. \hyperref[sec:appendix:add_qual]{Additional Quantitative Results}}
\dotfill \pageref{sec:appendix:add_qual}

\vspace{0.2em}
\hspace{2.0em}
\hyperref[app:3b_result_table]{C.1 \ Performance Breakdown for the 3B Student Model}
\dotfill \pageref{app:3b_result_table} 

\hspace{2.0em}
\hyperref[app:performance_vs_subset_size]{C.2 \ Performance on Different Subset Size}
\dotfill \pageref{app:performance_vs_subset_size} 

\vspace{0.2em}
\rule{\linewidth}{0.2pt}
\vspace{0.2em}

\noindent
\textbf{D. \hyperref[sec:app-full-opt-prompts]{Full Optimized Prompts}}
\dotfill \pageref{sec:app-full-opt-prompts}

\vspace{0.2em}
\hspace{2.0em}
\hyperref[sec:appendix:qwen25_3]{D.1 \ Qwen2.5-1.5B-Instruct}
\dotfill \pageref{sec:appendix:qwen25_3} 

\hspace{3.8em}
{\normalsize\hyperref[sec:appendix:sym_3]{D.1.1 \ \ourmethod}}
\dotfill \pageref{sec:appendix:sym_3} 

\hspace{3.8em}
{\normalsize\hyperref[sec:appendix:nl_3]{D.1.2 \ Opt-NL}}
\dotfill \pageref{sec:appendix:nl_3} 

\hspace{2.0em}
\hyperref[sec:appendix:qwen25]{D.2 \ Qwen2.5-3B-Instruct}
\dotfill \pageref{sec:appendix:qwen25} 

\hspace{3.8em}
{\normalsize\hyperref[sec:appendix:sym]{D.2.1 \ \ourmethod}}
\dotfill \pageref{sec:appendix:sym} 

\hspace{3.8em}
{\normalsize\hyperref[sec:appendix:nl]{D.2.2 \ Opt-NL}}
\dotfill \pageref{sec:appendix:nl}

\vspace{0.2em}
\rule{\linewidth}{0.2pt}
\vspace{0.2em}

\noindent

\textbf{E. \hyperref[app:qualititative-results]{Qualitative Results}}
\dotfill \pageref{app:qualititative-results}

\hspace{2.0em}
\hyperref[app:generated-problem-variations]{E.1 Examples of Generated Problems}
\dotfill \pageref{app:generated-problem-variations} 

\vspace{0.2em}

\hspace{3.8em}
{\normalsize\hyperref[sec:appendix:generated_problem_opt_nl]{E.1.1 \ Problems Generated with Opt-NL}}
\dotfill \pageref{sec:appendix:generated_problem_opt_nl}

\hspace{3.8em}
{\normalsize\hyperref[sec:appendix:generated_problem_our_method]{E.1.2 \ Problems Generated with \ourmethod}}
\dotfill \pageref{sec:appendix:generated_problem_our_method}

\hspace{2.0em}
{\normalsize\hyperref[app:representation-comparison]{E.2 Comparison of Problems Generated via Different Representations}}
\dotfill \pageref{app:representation-comparison}}

\newpage

\section{Implementation Details}
\label{sec:appendix:detail}

\subsection{RLVR}\label{app:rlvr-details}
\cref{tab:impl_details} shows the hyperparameters used for training the student model.

\begin{table}[h!]
\centering
\caption{{\bf Full hyperparameter setting.}}
\label{tab:impl_details}
\resizebox{0.4\linewidth}{!}{
\begin{tabular}{l|c|c}
\toprule
\textbf{Hyperparameters} & \textbf{GRPO} & \textbf{PPO} \\
\midrule
Training batch size        & 1024 & 1024 \\
Samples per prompt         & 8     & - \\
Max response length        & 1024  & 1024 \\
Clip ratio                 & 0.2   & 0.2 \\
Scaling factor $\alpha$    & 1.0   & - \\
\midrule
Training temperature       & 1.0   & 1.0 \\
Training top\_p            & 1.0   & 1.0 \\
Validation temperature     & 1.0   & 1.0 \\
Validation top\_p          & 1.0   & 1.0 \\
\midrule
Total gradient steps       & 200   & 200 \\
\midrule
Optimizer                  & AdamW & AdamW \\
Learning rate              & $1\times 10^{-6}$ & $1\times 10^{-6}$ \\
LR warmup steps            & 0     & 0 \\
LR scheduler               & constant & constant \\
\midrule
Critic model & -- & Qwen2.5-1.5B-Instruct \\
\bottomrule
\end{tabular}
}
\end{table}

\subsection{Prompt Optimization}
\label{app:prompt-opt-details}

\textbf{Loss.} As mentioned in~\cref{sec:closed-loop}, our loss for prompt optimization consists of a difficulty criterion $\mathcal{C}_\text{diff}$ and representation-specific guardrails $\mathcal{C}_\text{guard}$.

\textbf{Difficulty Criterion.} The base difficulty criterion is:

$\mathcal{C}_\text{diff} =$ \texttt{\small Judge mathematical difficulty, not verbosity. The problem should be challenging - the student should get the wrong answer.}

To introduce variance in our optimized prompts, we define four variations of $\mathcal{C}_\text{diff}$ by appending additional instructions. These variations are applied during prompt optimization for both NL and Sym representations:

\begin{itemize}[leftmargin=*, nosep]
    \item $\mathcal{C}_\text{diff}^{(0)} = \mathcal{C}_\text{diff}$ (base criterion only)
    \item $\mathcal{C}_\text{diff}^{(1)}$ extends $\mathcal{C}_\text{diff}$ with: {\small\texttt{Increase problem difficulty through local changes (e.g., altering numbers, variable names, or relationships).}}
    \item $\mathcal{C}_\text{diff}^{(2)}$ extends $\mathcal{C}_\text{diff}$ with: {\small\texttt{The modified code should include additional algebraic transformations or nested symbolic expressions that require multi-step simplification before the final answer.}}
    \item $\mathcal{C}_\text{diff}^{(3)}$ extends $\mathcal{C}_\text{diff}$ with: {\small\texttt{The modified code should introduce new interdependent quantities or chained sub-calculations that must be combined algebraically to obtain the final answer.}}
\end{itemize}

\textbf{Guardrails.} We define representation-specific guardrails to ensure quality and feasibility:

For natural language problems ($\mathcal{C}_\text{guard}^\text{NL}$):
\begin{lstlisting}
Clarity - Simple, unambiguous wording; all quantities and units explicit.
Length - Concise; at most 50% longer than original
\end{lstlisting}

For symbolic code problems ($\mathcal{C}_\text{guard}^\text{Sym}$):
\begin{lstlisting}
Cleanliness - Code can be easily translated into a word problem. Penalize added boilerplate, comments, or helper symbols that bloats the code length.    
\end{lstlisting}

For \ourmethod, we additionally enforce programmatic checks on the modified SymPy code to ensure: (1) it has a unique answer, (2) it runs without errors, and (3) its length is at most 50\% longer than the original.

\subsection{Data Generation}\label{app:data-generation}

The prompts used for getting the solution from a word problem and for translating the symbolic representation to a word problem is shown below.

$p_\text{solve}=$ \texttt{\small Given this problem, reason step by step, and put your final answer within boxed\{\{\}\}. Return the exact answer without rounding or approximation.}

$p_\text{render}=$  \texttt{\small Translate the following code into a real-world style math word problem. 
Use the values, operations, and relationships in the code to construct a coherent scenario. 
Make sure each variable and equation is represented naturally in the problem's narrative. Keep the problem realistic and readable. The problem should end with exactly one question that asks for the unknown quantity represented by the variable named `answer'. Do not mention the variable named \'answer\' explicitly in the text. Return only the word problem and nothing else.}
\section{Qualitative Analysis}\label{app:qualitative-analysis}

We present a concrete example illustrating how a simple arithmetic word problem is progressively
enriched into a more structured reasoning task, and finally transformed into an explicit algebraic
constraint-based formulation. The original problem and the problems transformed by Opt-NL and \ourmethod are shown in full,
with linguistic components visually aligned to their corresponding mathematical structures.

\vspace{1em}

\noindent
\textbf{Original Problem}
\begin{quote}
Natalia sold clips to \textcolor{Preserve}{\textbf{48}} of her friends in April, and then she sold
\textcolor{Implicit}{\textbf{half as many clips}} in May.
How many clips did Natalia sell altogether in April and May?
\end{quote}

\noindent
\textbf{Implied Computation (Original)}
\[
\textcolor{Preserve}{48}
\;+\;
\textcolor{Implicit}{\frac{1}{2} \cdot 48}
\]

\vspace{0.5em}
\hrule
\vspace{0.5em}

\noindent
\textbf{Opt-NL Transformed Problem}
\begin{quote}
Natalia sold clips to \textcolor{Preserve}{\textbf{48 different friends}} in April; each friend bought
either \textcolor{Implicit}{\textbf{1 clip or 2 clips}}.
In May, she sold \textcolor{Preserve}{\textbf{half as many clips as she did in April}}.
The number of friends who bought \textcolor{Injected}{\textbf{2 clips in April}}
equals \textcolor{Injected}{\textbf{two-thirds of the number of clips sold in May}}.
How many clips did Natalia sell altogether in April and May?
\end{quote}

\noindent
\textbf{Implicit Mathematical Structure (Transformed)}
\[
\begin{aligned}
\textcolor{Preserve}{a + b} &= \textcolor{Preserve}{48}
&& \text{(number of friends)} \\[0.3em]
\textcolor{Implicit}{\text{April clips}} &= a + 2b
&& \text{(clip count model)} \\[0.3em]
\textcolor{Preserve}{\text{May clips}} &= \textcolor{Preserve}{\frac{1}{2}(a + 2b)}
&& \text{(preserved relation)} \\[0.3em]
\textcolor{Injected}{b} &= \textcolor{Injected}{\frac{2}{3} \cdot \text{May clips}}
&& \text{(additional constraint)}
\end{aligned}
\]

\vspace{0.5em}
\hrule
\vspace{0.5em}

\noindent
\textbf{\ourmethod Transformed Problem}
\begin{quote}
A warehouse has two storage tanks whose capacities
\textcolor{Preserve}{\textbf{add up to 48 liters}}.
The capacity of the larger tank
\textcolor{Injected}{\textbf{times the capacity of the medium tank is 512 liters squared}},
and the \textcolor{Ordering}{\textbf{larger tank is bigger than the medium one}}.
After transferring
\textcolor{Implicit}{\textbf{half of the medium tank’s contents into the larger tank}},
how many liters are in the larger tank?
\end{quote}

\noindent
\textbf{Explicit Algebraic Formulation (Transformed)}
\[
\begin{aligned}
\textcolor{Preserve}{x + y} &= \textcolor{Preserve}{48}
&& \text{(preserved total)} \\[0.3em]
\textcolor{Injected}{x \cdot y} &= \textcolor{Injected}{512}
&& \text{(nonlinear constraint)} \\[0.3em]
\textcolor{Ordering}{x} &> \textcolor{Ordering}{y}
&& \text{(ordering constraint)} \\[0.3em]
\textcolor{Implicit}{x'} &= x + \textcolor{Implicit}{\frac{1}{2}y}
&& \text{(explicit state transition)}
\end{aligned}
\]

\vspace{0.5em}

\paragraph{Why This Progression Matters.}
The original problem requires only a single arithmetic operation with one implicit quantity.
The enriched word problem introduces latent variables and relational constraints while remaining
in natural language. The transformed problem makes all constraints explicit and symbolic,
requiring the solver to identify a valid initial state from a system of equations before
applying a state transition. This progression highlights how surface-level arithmetic reasoning
can be elevated into structured algebraic modeling.


\noindent
\textbf{Original Problem}
\begin{quote}
A deep-sea monster rises from the waters once every
\textcolor{Preserve}{\textbf{hundred years}} to feast on a ship.
Over
\textcolor{Preserve}{\textbf{three hundred years}},
it has consumed
\textcolor{Preserve}{\textbf{847 people}}.
Ships have been built larger over time, so each new ship has
\textcolor{Implicit}{\textbf{twice as many people as the last ship}}.
How many people were on the ship the monster ate in the first hundred years?
\end{quote}

\noindent
\textbf{Implied Computation (Original)}
\[
x + 2x + 4x = \textcolor{Preserve}{847}
\]

\vspace{0.5em}
\hrule
\vspace{0.5em}

\noindent
\textbf{Opt-NL Transformed Problem}
\begin{quote}
A deep-sea monster rises from the waters once every
\textcolor{Preserve}{\textbf{100 years}} to feast on a ship.
Over
\textcolor{Preserve}{\textbf{300 years}} it has consumed
\textcolor{Preserve}{\textbf{847 people}}.
Each new ship’s intended number of people is
\textcolor{Implicit}{\textbf{twice as many as on the previous ship}}.
However, before the second voyage
\textcolor{Injected}{\textbf{17 intended passengers were lost ashore}},
and before the third voyage
\textcolor{Injected}{\textbf{3 extra passengers joined for a festival}}.
How many people were on the ship the monster ate in the first hundred years?
\end{quote}

\noindent
\textbf{Implicit Mathematical Structure (Enriched)}
\[
\begin{aligned}
\textcolor{Preserve}{\text{First ship}} &= x \\[0.3em]
\textcolor{Implicit}{\text{Second ship}} &= 2x - \textcolor{Injected}{17} \\[0.3em]
\textcolor{Implicit}{\text{Third ship}} &= 4x + \textcolor{Injected}{3} \\[0.3em]
\textcolor{Preserve}{x + (2x - 17) + (4x + 3)} &= \textcolor{Preserve}{847}
\end{aligned}
\]

\vspace{0.5em}
\hrule
\vspace{0.5em}

\noindent
\textbf{\ourmethod Transformed Problem}
\begin{quote}
Three craft workshops run three specialized machines that produce identical widgets.
The slowest machine makes a whole number of widgets per hour; call its rate
\textcolor{Preserve}{\textbf{$a$}}.
The second machine makes
\textcolor{Implicit}{\textbf{twice as many widgets per hour as the first minus one}},
and the
\textcolor{Injected}{\textbf{product of their hourly rates equals 120}}.
The third machine’s hourly rate equals
\textcolor{Injected}{\textbf{the sum of the first two rates minus six}}.
If each machine runs continuously for
\textcolor{Preserve}{\textbf{three hours}},
how many widgets do all three machines produce in total?
\end{quote}

\noindent
\textbf{Explicit Algebraic Formulation (Transformed)}
\[
\begin{aligned}
\textcolor{Preserve}{a} &\in \mathbb{Z}_{>0}
&& \text{(base rate)} \\[0.3em]
\textcolor{Implicit}{b} &= \textcolor{Implicit}{2a - 1}
&& \text{(derived rate)} \\[0.3em]
\textcolor{Injected}{a \cdot b} &= \textcolor{Injected}{120}
&& \text{(nonlinear constraint)} \\[0.3em]
\textcolor{Injected}{c} &= \textcolor{Injected}{a + b - 6}
&& \text{(additive dependency)} \\[0.3em]
\textcolor{Preserve}{\text{Total output}} &= 3(a + b + c)
&& \text{(state accumulation)}
\end{aligned}
\]

\vspace{0.5em}

\paragraph{Why the Transformation Is Stronger.}
The original problem relies on recognizing a geometric progression across discrete time steps.
The enriched version introduces asymmetric perturbations that break the pure doubling pattern while preserving a solvable linear structure. The transformed question removes the temporal and narrative framing entirely and replaces it with a system of interacting nonlinear and linear constraints, requiring explicit variable definition, constraint satisfaction, and final aggregation.
This reframing elevates the task from pattern recognition to formal algebraic modeling.


\section{Additional Quantitative Results}
\label{sec:appendix:add_qual}

\subsection{Performance breakdown for the 3B model}
\label{app:3b_result_table}

\cref{tab:results_main_3b} shows the breakdown of performance for the 3B model in the same format as for the 1.5B model in the main paper.

\begin{table*}[h]
\centering
\caption{Accuracies on several benchmarks before and after training the 3B student model with different types of generated data and RL methods.}
\begin{adjustbox}{max width=\linewidth}
\begin{tabular}{ll>{\columncolor{lightgray}}c llllllll}
\toprule
\textbf{RL Method} & \textbf{Data type} & \textbf{Average} & \textbf{GSM8K} & \textbf{GSM-Sym} & \textbf{Sym-p1} & \textbf{Sym-p2} & \textbf{GSM-Plus} & \textbf{MATH-500} & \textbf{AIME24} \\
\midrule

 & None
 &59.35 &85.37 &80.8 &69.34 &45.04 &62.5	 &62.37 &10\\
\midrule

\multirow{6}{*}{\textbf{PPO}}
 & Seed-data    & 60.47 & 87.04 & 85.04 & 74.40 & 49.60 & 67.25 & 59.96 & 0.00 \\
 & Base-NL      & 62.22 & 87.04 & 84.82 & 75.64 & 50.88 & 68.54 & 61.97 & \textbf{6.67} \\
 & Base-Sym     & 61.07 & 87.34 & \textbf{85.24} & 74.96 & 48.72 & 68.38 & 59.56 & 3.33 \\
 & Opt-NL       & \textbf{63.10} & \textbf{87.87} & 84.72 & \textbf{76.98} & 52.24 & \textbf{69.25} & \textbf{63.98} & \textbf{6.67} \\
 & \ourmethod   & 62.50 & 87.64 & 84.64 & 75.82 & \textbf{53.12} & 67.67 & 61.97 & \textbf{6.67} \\
\cmidrule(lr){2-10}
 & \textbf{$\Delta$ (vs. Seed)}
 & \textbf{\pos{+2.03}} & \textbf{\pos{+0.60}} & \textbf{\pos{-0.40}} & \textbf{\pos{+1.42}}
 & \textbf{\pos{+3.52}} & \textbf{\pos{+0.42}} & \textbf{\pos{+2.01}} & \textbf{\pos{+6.67}} \\
\midrule

\multirow{6}{*}{\textbf{GRPO}}
 & Seed-data    & 59.35 & 85.37 & 80.80 & 69.34 & 45.04 & 62.50 & 62.37 & 10.00 \\
 & Base-NL      & 63.36 & \textbf{87.95} & 85.08 & 76.44 & 51.80 & \textbf{69.29} & 62.98 & 10.00 \\
 & Base-Sym     & 62.02 & 86.88 & \textbf{85.70} & 73.88 & 53.76 & 67.67 & 59.56 & 6.67 \\
 & Opt-NL       & 62.57 & 87.41 & 85.14 & 76.10 & 53.20 & 68.71 & 60.76 & 6.67 \\
 & \ourmethod   & \textbf{64.39} & 87.64 & 84.34 & \textbf{76.95} & \textbf{56.00} & 68.92 & \textbf{63.58} & \textbf{13.33} \\
\cmidrule(lr){2-10}
 & \textbf{$\Delta$ (vs. Seed)}
 & \textbf{\pos{+5.04}} & \textbf{\pos{+2.27}} & \textbf{\pos{+3.54}} & \textbf{\pos{+7.56}}
 & \textbf{\pos{+10.96}} & \textbf{\pos{+6.42}} & \textbf{\pos{+1.21}} & \textbf{\pos{+3.33}} \\

\bottomrule
\end{tabular}
\end{adjustbox}
\label{tab:results_main_3b}
\end{table*}

\subsection{Performance on Different Subset Size}
\label{app:performance_vs_subset_size}

\cref{fig:full_subset_results} shows the performance from training on different amounts of generated data for the 1.5B and 3B student model. For the 1.5B model, we show this for both GRPO and PPO, and when trained on filtered (i.e., using GPT-5-mini to annotate if the generated problem has the correct solution, if it does not, replace that datapoint with the baseline generated data), and unfiltered data. 

\begin{figure}[!h]
    \centering
    \includegraphics[width=0.98\linewidth]{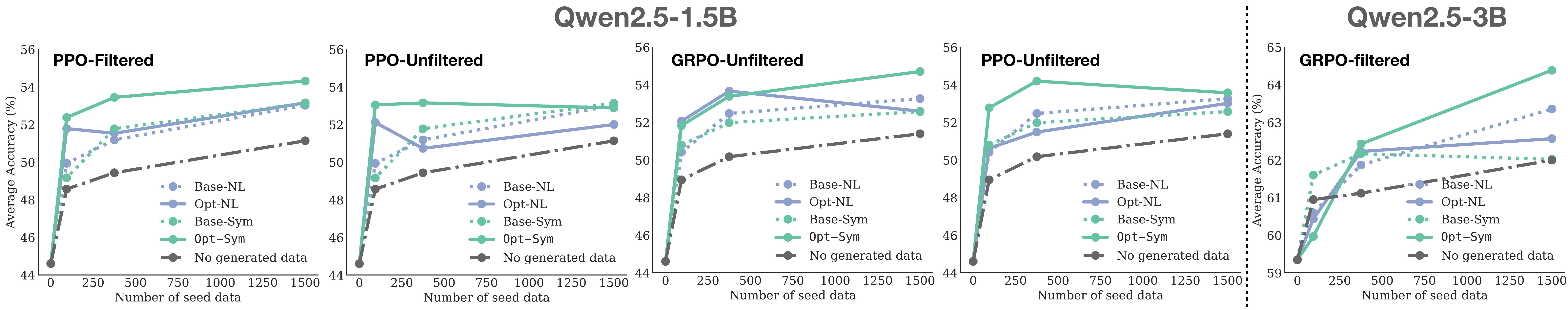}
    \caption{Performance vs different subsets of data for different RL methods and student models.}
    \label{fig:full_subset_results}
\end{figure}

\section{Full optimized prompts}
\label{sec:app-full-opt-prompts}

\subsection{For Qwen2.5-1.5B-Instruct}
\label{sec:appendix:qwen25_3}

\subsubsection{\ourmethod}
\label{sec:appendix:sym_3}

\begin{promptboxSYM}{1}
\ttfamily\small
Modify this SymPy code to increase mathematical difficulty by introducing a concise, multi-step algebraic problem that maintains the original format. Use meaningful, context-linked variable names with explicit domain assumptions (e.g., positive integers). Structure the code into clear sections with minimal but precise comments explaining the real-world meaning of variables and equations. Incorporate at least one nonlinear or chained constraint (such as a product, ratio, or sum-of-squares) that deepens algebraic reasoning without overcomplicating or contradicting other constraints. 
Prefer explicit integer solution search or enumeration over purely symbolic solving to ensure integrality, positivity, and uniqueness of solutions. Parameterize constants with example values that guarantee a unique, physically plausible numeric solution. Include a final step that fully simplifies and numerically evaluates the unique answer to a single integer or rational number, assigning it to a clearly named variable (e.g., `answer`). Avoid placeholders or vague comments; ensure all comments are informative and directly tied to the code. Optionally modularize the code into logical blocks (variable definitions, equations, solving, filtering, final answer) with section headers to facilitate stepwise reasoning and natural language conversion. Include a minimal inline test or assertion confirming the uniqueness and validity of the solution. Ensure the entire code snippet is syntactically complete, concise, and executable as-is, suitable for conversion into a challenging math word problem that encourages algebraic reasoning without changing the problem’s format.
\end{promptboxSYM}

\begin{promptboxSYM}{2}
\ttfamily\small
Modify this SymPy code snippet to increase mathematical difficulty by introducing fully defined, meaningful, context-rich symbolic variables that represent real-world quantities. Construct multi-step symbolic expressions involving nested rational functions, polynomial terms, and composite operations (such as sums inside products or rational coefficients with differing denominators) that require genuine algebraic manipulation—like expansion, factoring using algebraic identities (e.g., difference of squares, perfect square trinomials), and simplification—before any numeric substitution. 
Maintain symbolic reasoning throughout intermediate steps, deferring numeric evaluation until the final step. Ensure all variables and expressions are explicitly defined and logically connected to form a coherent, multi-step problem narrative. Assign the final numeric answer to a clearly named variable (e.g., `answer`) derived from fully defined symbolic expressions, verifying it is a unique, fully simplified SymPy Integer or Rational object (not symbolic or a list). Include minimal, precise inline comments that clarify the mathematical purpose of each step without clutter, facilitating natural language conversion. Avoid undefined or unused variables, redundant imports, placeholders, or ellipses. Incorporate domain restrictions or assumptions on symbolic variables to guarantee solution validity and uniqueness, and include explicit checks or assertions confirming exactly one valid numeric solution exists before assigning the final answer. Structure the code to mirror logical problem-solving steps, enhancing clarity and interpretability. Use exact rational arithmetic throughout, and combine algebraic operations where appropriate to maintain concise, clean, and error-free code suitable for generating challenging, fair, and solvable natural language math word problems.
\end{promptboxSYM}

\begin{promptboxSYM}{3}
\ttfamily\small
Make small, local algebraic edits that increase mathematical difficulty by introducing rational coefficients, fractional multipliers, and explicit intermediate variables with clear, descriptive names. 
Use SymPy’s `sp.Rational` for all fractional values to maintain exact arithmetic, and replace any floor division `//` with `sp.floor()` for integer truncation. Define all intermediate variables fully and explicitly, such as `subtotal` as the sum of item costs, `tax\_amount` as tax on subtotal, and `discount\_amount` as a function of integer division of quantities. Maintain consistent snake\_case variable naming reflecting real-world entities (e.g., `num\_pizza\_boxes`, `price\_per\_drink\_box`, `pre\_tax\_total`, `final\_total`). Avoid unnecessary wrapping of integers with `sp.Integer()` unless symbolically required. Structure calculations to clearly model the order of operations: compute subtotal, then tax, then discount, then final total. Include brief inline comments explaining each step to aid interpretability. Ensure the final `answer` variable is assigned a fully simplified, unique numeric value (integer or rational) using `sp.simplify()`. Keep code concise and within format constraints, avoiding redundant variables or overly complex numeric wrapping. Optionally, introduce symbolic parameters assigned rational values to enable flexible difficulty tuning without changing the problem format.
\end{promptboxSYM}

\begin{promptboxSYM}{4}
\ttfamily\small
Modify this SymPy code to increase mathematical difficulty by introducing algebraic complexity such as nonlinear terms or quadratic summations, while strictly preserving the original problem format and generator rails. 
Ensure all variables are explicitly defined with meaningful, context-reflective names and that all expressions are complete and syntactically correct. Replace symbolic functions like trigonometric terms with their exact numeric equivalents (e.g., use sp.Rational(1, 2) instead of sp.sin(sp.pi/6)) before final simplification to guarantee a unique numeric final answer. Use closed-form expressions for summations (e.g., sum of squares formula) instead of symbolic summation notation to improve clarity and facilitate natural language conversion. 
Add minimal inline comments describing the role of each variable and calculation step to guide downstream natural language generation without altering the format. Balance complexity to avoid overloading the problem—consider linearizing nonlinear components or simplifying trigonometric tweaks if needed. Explicitly compute and assign the final answer as a single integer or rational number using SymPy’s evaluation methods (.evalf(), .doit(), or direct substitution). Validate that the modified code runs without errors and that the final answer is a unique numeric value. Avoid undefined variables or placeholders and ensure all intermediate variables are fully specified. Maintain concise, clean, and well-structured code that facilitates smooth conversion into a natural language math word problem. Do NOT include any introductory phrases, meta-questions, evaluations, or rubrics.
\end{promptboxSYM}

\subsubsection{Opt-NL}
\label{sec:appendix:nl_3}

\begin{promptboxNL}{1}
\ttfamily\small
Modify this problem to increase mathematical difficulty by introducing algebraic interdependencies or chained calculations that require multi-step reasoning and solving for variables, without changing the problem format. Embed elements that commonly cause student errors or misconceptions (such as ambiguous references, fractional parts, or multi-step proportional reasoning) to test understanding. Ensure the problem remains clear, unambiguous, and fair, avoiding excessive cognitive load or confusing phrasing. Retain explicit units and clearly define all quantities. Maintain or enhance the problem’s relatable context to keep it engaging and accessible. Limit the problem length increase to no more than 50\% of the original. Internally self-check the problem for logical consistency, solvability, and clarity before outputting. Return ONLY the revised problem text ending with exactly one question that asks for a single numerical answer. Do NOT include any introductory phrases, meta-questions, evaluations, or rubrics.
\end{promptboxNL}

\begin{promptboxNL}{2}
\ttfamily\small
Modify this problem to increase mathematical difficulty by introducing multi-step algebraic or symbolic expressions that require nested simplification or solving for variables. Ensure the revised problem maintains the original format and context, remains clear and unambiguous, and is no more than 50\% longer than the original. Design the problem to provoke common student errors such as misapplying algebraic transformations or misinterpreting nested expressions, while keeping it fair and solvable. Return ONLY the revised problem text ending with exactly one question that asks for a single numerical answer. Do NOT include any introductory phrases, meta-questions, evaluations, or rubrics.
\end{promptboxNL}

\begin{promptboxNL}{3}
\ttfamily\small
Make small, local algebraic edits that meaningfully increase mathematical difficulty by introducing more complex fractions (e.g., 3/7, 5/8) or mixed numbers, fractional or non-integer percentages (e.g., 12.5\%), and multi-step calculations that require intermediate reasoning without changing the problem’s format or length by more than 50\%. Ensure edits target common student error patterns—such as confusing fractions and percentages, misapplying percentage increases to subsets, or mixing total and partial quantities—while maintaining clear, concise wording and a single, unambiguous numerical-answer question. Prefer algebraic modifications that introduce variables, simple expressions, or conditional relationships to deepen reasoning (e.g., requiring solving for unknowns or comparing quantities algebraically) without adding extraneous text. Internally verify that the revised problem remains solvable, fair, and that the difficulty increase is likely to cause typical student errors, avoiding trivial numeric scaling or excessive complexity. Keep all edits domain-appropriate, focused on algebraic operations like fractions, percentages, multiplication, and addition, and preserve the original problem’s style and context.
\end{promptboxNL}

\begin{promptboxNL}{4}
\ttfamily\small
Modify this problem to increase mathematical difficulty by introducing additional variables, relationships, or constraints that require multi-step algebraic reasoning, while ensuring the problem remains clear, fair, and solvable. Maintain or improve clarity by explicitly stating all new relationships and avoiding ambiguous or implicit assumptions. Keep the problem concise, avoiding redundant or overly wordy phrasing, and do not increase the problem length by more than 50\%. Preserve the original problem’s context and ensure any new elements are age-appropriate and relevant. Incorporate subtle elements that encourage careful interpretation and common student pitfalls without causing undue frustration. Before finalizing, verify that the problem can be solved with the given information and that it ends with exactly one clear, unambiguous question asking for a single numerical answer. Do NOT include any introductory phrases, meta-questions, evaluations, or rubrics in the output.
\end{promptboxNL}

\subsubsection{SMT-LIB}

\begin{promptboxSMT}{1}
\ttfamily\small
Modify this SMT-LIB program to increase or decrease problem difficulty by introducing meaningful constraints or variable dependencies while preserving the original format and ensuring exactly one variable is queried in (get-value (...)).Replace some fixed constants with variables constrained by equalities or inequalities to create nontrivial reasoning challenges. Add domain constraints such as non-negativity or integer restrictions where appropriate. Introduce subtle integer division or remainder conditions to enhance complexity without changing problem structure. Remove redundant declarations or assertions to maintain conciseness and clarity. Embed minimal SMT-LIB comments (using `;`) within the code to clarify variable roles, domain assumptions, or difficulty adjustments without breaking format.Append a fixed-format comment block after the SMT-LIB code (if allowed) that briefly confirms SMT syntax correctness, uniqueness of solution, and the nature of difficulty adjustments made.Return ONLY the revised SMT-LIB code with these enhancements, ending with exactly one answer (one variable in (get-value (...))). Do NOT include any introductory phrases, meta-questions, or evaluations outside the SMT-LIB code and allowed comment block.
\end{promptboxSMT}

\begin{promptboxSMT}{2}
\ttfamily\small
Modify the given math problem by introducing one or two subtle, conceptually rich changes that increase its difficulty without altering the original format, style, or length by more than 30\%. Ensure the modifications embed common, identifiable student misconceptions related to the new elements (e.g., ignoring a discount, misapplying fractions, or miscounting participants) to make the difficulty meaningful and diagnostic. Internally verify that the revised problem remains clear, fair, and solvable, with unambiguous phrasing especially for any conditional or piecewise clauses students can correctly interpret and apply the new complexity without guesswork. Frame the final question to highlight the key challenging aspect introduced, possibly by including a minimal, natural contextual cue or hint that nudges students toward the intended reasoning path without giving away the solution. Maintain consistent numerical formats and explicitly specify or imply the expected answer format and units to avoid ambiguity. Do not include any calculations, solution steps, evaluative commentary, or meta-text output only the revised problem text. Before outputting, internally confirm that the problem complexity aligns with the intended difficulty increase, that the new element is central to evaluating student correctness, and that the problem preserves the original question style and single numerical answer format.

\end{promptboxSMT}

\begin{promptboxSMT}{3}
\ttfamily\small
Modify the SMT-LIB program to increase its difficulty by making carefully considered, nontrivial adjustments to numeric constants and arithmetic relationships that preserve the original SMT-LIB format, variable declarations, and single (get-value (...)) query. Ensure that these changes maintain or enhance the conceptual complexity—such as introducing meaningful multiplicative versus additive distinctions, nested arithmetic expressions, and logical dependencies—without altering the logical flow or structure of constraints. Preserve the problem’s satisfiability and guarantee a unique solution for the queried variable by internally verifying consistency and uniqueness. Select numeric values that create pedagogically relevant contrasts (e.g., factors slightly less than 1 to simulate discounts or offsets that highlight conceptual differences) while respecting SMT integer semantics, including the effects of (to\_int ...) truncation and (div ...) integer division. Maintain alignment between the SMT problem and its real-world interpretation, avoiding fractional or nonsensical solutions. Preserve variable types and roles, and avoid introducing redundant variables or assertions. If SMT-LIB syntax permits, include minimal inline comments to annotate key conceptual points or logical steps to aid downstream evaluation; otherwise, structure assertions clearly and logically. Balance difficulty increase with accessibility by making moderate, realistic numeric adjustments that enhance problem interpretability and relevance. Finally, ensure the modified SMT-LIB code is syntactically valid, semantically correct, and ready for immediate evaluation without any introductory or evaluative text beyond the SMT code itself.
\end{promptboxSMT}

\begin{promptboxSMT}{4}
\ttfamily\small
Modify this SMT-LIB program to increase its reasoning complexity by embedding up to two subtle, logically meaningful constraints or variables that preserve the original problem’s format and output structure. Specifically:

1. Introduce new variables or constraints only if they meaningfully contribute to the problem’s logic and increase cognitive load through multi-layered or interdependent conditions (e.g., nested `ite` expressions, modular arithmetic combined with inequalities, or threshold-based piecewise definitions).

2. Prefer minimal, canonical forms of constraints—such as defining adjustment variables directly with `(mod ...)` rather than verbose or redundant `ite` expressions—to maintain solver efficiency and human readability.

3. Explicitly constrain any new variables’ domains (e.g., binary variables constrained to 0 or 1) to avoid ambiguity and unintended interpretations.

4. Adjust numeric parameters subtly within ±15\%, choosing values that create non-trivial cases (e.g., non-integer divisions or borderline thresholds) and include concise inline comments explaining the rationale behind these changes to enhance transparency.

5. Add clear, descriptive inline SMT-LIB comments adjacent to all new variables and assertions, explaining their purpose, how they increase difficulty, and how they relate to the original problem or common student reasoning errors—especially clarifying integer division semantics or rounding behavior.

6. Ensure comments accurately reflect the implemented logic (e.g., if an adjustment rounds up, the comment should state so explicitly) to reduce confusion and improve interpretability.

7. Maintain a clear, stepwise reasoning pipeline by defining intermediate variables where appropriate, grouping related assertions logically, and preserving the original problem’s domain and educational level.

8. Include explicit assertions or comments that confirm the uniqueness and satisfiability of the solution after modifications, reinforcing problem validity and helping downstream users trust the solution.

9. Avoid redundant or conflicting assertions and unnecessary variables that do not influence the final queried value, ensuring a clean, focused encoding.

10. Enforce consistent formatting, indentation, and spacing throughout the SMT-LIB code to improve readability.

11. Preserve the problem’s length within ±15\% and maintain exactly one `(get-value (...))` query, which should correspond to the variable representing the problem’s final answer.

By following these guidelines, produce a revised SMT-LIB program that subtly increases reasoning complexity, embeds pedagogically meaningful difficulty, and remains clear, precise, and solver-friendly without altering the problem’s original format or output structure.
\end{promptboxSMT}

\subsection{Qwen2.5-3B-Instruct}
\label{sec:appendix:qwen25}

\subsubsection{\ourmethod}
\label{sec:appendix:sym}

\begin{promptboxSYM}{1}
\ttfamily\small
Make small, local algebraic edits that increase symbolic complexity and cognitive challenge by introducing non-integer rational coefficients with varied denominators, non-perfect square radicals that do not trivially cancel, and combined fractional terms. Preserve the original code format and variable naming conventions but allow minimal, contextually meaningful renaming or aliasing to enhance semantic clarity. Decompose complex expressions into intermediate symbolic variables with descriptive names and concise inline comments that clarify their mathematical or real-world roles, supporting traceability and natural language problem generation. Represent all constants and coefficients exactly using sp.Rational or symbolic expressions to maintain precision and avoid floating-point approximations. Ensure every variable used in the final computation is explicitly defined within the snippet, removing unused or redundant code fragments. Compute the final answer explicitly as a symbolic expression derived from these variables, delaying simplification until the end and applying sp.simplify() or sp.cancel() to produce a unique, fully simplified numeric value. Optionally, include symbolic parameters to enable problem parameterization and reusability. Structure the code into logical blocks reflecting algebraic steps, and embed minimal but meaningful comments to guide downstream natural language generation without increasing code length unnecessarily. Avoid trivial zero-value expressions or canceling terms that add no genuine complexity. Balance increased symbolic complexity with readability and pedagogical naturalness, ensuring the code remains concise, error-free, and suitable for generating coherent, cognitively challenging math word problems.
\end{promptboxSYM}

\begin{promptboxSYM}{2}
\ttfamily\small
Modify this SymPy code to increase mathematical difficulty by introducing a nonlinear equation with multiple roots and explicit symbolic domain constraints that ensure the final answer is uniquely determined and numerically evaluable. Specifically, define symbolic variables with assumptions (e.g., positivity, integrality), construct a nonlinear polynomial equation (such as a cubic or quartic) involving these variables, and expand it fully to expose the polynomial form. Incorporate symbolic inequalities to represent domain restrictions and use SymPy's solving and inequality-solving functions to find all candidate roots. Filter these roots symbolically and numerically to retain only those satisfying all domain constraints, discarding extraneous solutions arising from nonlinear transformations. Maintain symbolic expressions throughout the solving process, substituting numeric values only at the final step to compute a unique, simplified numeric answer assigned to the variable answer. Use clear, meaningful variable names and concise comments that explain the role of each variable, the nonlinear constraint, and the domain restrictions in plain mathematical terms, facilitating natural language problem generation. Ensure the code is fully self-contained, syntactically correct, and executable as-is, producing a unique numeric final answer without errors. Include a final assertion or check confirming the uniqueness and validity of the solution. For example, define variables r and k with domain constraints, form and expand the cubic polynomial equation 90*(1 + r)*(1 + r + k*r**2) - 216 = 0, solve symbolically for r, filter roots by positivity and domain inequalities, and assign the unique valid numeric root to answer. This approach balances increased symbolic complexity with clarity and completeness, embedding genuine nonlinear difficulty that meaningfully affects the final numeric answer while respecting fixed problem format constraints.
\end{promptboxSYM}

\begin{promptboxSYM}{3}
\ttfamily\small
Make small, local algebraic edits that increase mathematical difficulty by introducing nontrivial symbolic expressions—such as factorials of 3 or 4, square roots of non-perfect squares (e.g., sqrt(2), sqrt(50)/5), or rational exponents that do not simplify immediately--while preserving the original code format and output assignment to answer. Ensure all variables are explicitly defined with meaningful, contextually clear names, and avoid circular or incomplete assignments. Introduce interdependent, multi-step algebraic expressions by defining intermediate variables that reflect real-world groupings or steps, thereby enriching the algebraic structure and facilitating natural language conversion. Refrain from premature simplification; keep symbolic expressions intact to encourage stepwise reasoning. Define percentages explicitly as rational numbers (e.g., sp.Rational(3,4)) or symbolic square roots of rational numbers that do not simplify trivially. Maintain concise, syntactically valid, and clean code without redundant operations. Where possible, add brief inline comments or meaningful variable names to implicitly encode problem context and support downstream natural language generation, but do NOT include any introductory phrases, meta-questions, evaluations, or rubrics.
\end{promptboxSYM}

\begin{promptboxSYM}{4}
\ttfamily\small
Modify this SymPy code to increase mathematical difficulty by introducing a multi-step symbolic solving process involving nonlinear or rational equations that require careful algebraic manipulation. Define multiple interdependent symbolic variables with clear, descriptive names reflecting real-world quantities, and include explicit assumptions (e.g., positivity, integrality) to model domain constraints and realistic conditions. Incorporate parameterized constants and subtle inequalities or piecewise conditions that reflect common student misconceptions or realistic problem constraints, ensuring these are fully documented with concise, context-rich comments linking variables and equations to their real-world meanings. Structure the code logically into sections: parameter definitions with assumptions, symbolic variable declarations, equation setup, symbolic solving producing a simplified symbolic formula for the unknown, followed by numeric substitution and evaluation yielding a unique, simplified numeric final answer assigned to a clearly named variable (e.g., answer). Use SymPy's simplification and numeric evaluation functions judiciously to maintain clarity and ensure uniqueness of the solution. Avoid redundant symbolic solves or unnecessary intermediate variables to keep the code concise (preferably within 8 lines) and clean. Include validation steps to confirm solution uniqueness and numeric finality, and ensure the final code snippet is fully self-contained and executable without placeholders or ambiguous indexing. This approach balances algebraic richness with readability and pedagogical value, producing a problem that is mathematically challenging, semantically clear, and well-suited for natural language math word problem generation.
\end{promptboxSYM}

\subsubsection{Opt-NL}
\label{sec:appendix:nl}

\begin{promptboxNL}{1}
\ttfamily\small
Make precise, local algebraic edits that increase the problem mathematical difficulty by introducing or enhancing interdependent quantities and chained calculations requiring multi-step reasoning. Ensure these edits: Embed algebraic relationships that necessitate sequential application of multiple percentage calculations with clear conditional inclusions or exclusions (e.g., some fees taxable, others not), thereby increasing cognitive load meaningfully. Explicitly define all new variables and units, clarifying their scope and role (e.g., one-time vs. per-item fees), and phrase all conditions unambiguously using parallel structure or restatements to minimize misinterpretation.Anticipate and incorporate algebraic structures that commonly cause student errors, such as distinguishing additive versus multiplicative relationships or conditional percentage applications, to create pedagogically valuable difficulty.- Maintain the original problem domain and context strictly, avoiding any introduction of unrelated, business, finance, or adult themes. Preserve the problem format and length, ensuring the revised problem length does not exceed 150 \% of the original, actively estimating length increase and avoiding unnecessary verbosity. Guarantee exactly one clear, unambiguous question that requires a single, straightforward numerical answer, and verify that the problem remains solvable with a verifiable final numeric value. Include concise clarifying phrases or sentences within the problem text as needed to explicitly state calculation sequences or dependencies, improving clarity without significantly increasing length.- Balance increased difficulty with fairness and accessibility, ensuring the complexity challenges students appropriately without overwhelming or confusing them.Avoid superficial numeric changes; instead, introduce genuine algebraic complexity that requires multi-step reasoning and variable interdependencies.- Internally review the revised problem for consistency, clarity, and solvability before finalizing, confirming that all algebraic relationships are logically coherent and that the question is unambiguous.Return only the revised problem text adhering to these guidelines
\end{promptboxNL}

\begin{promptboxNL}{2}
\ttfamily\small
Modify this problem to increase its mathematical difficulty by introducing meaningful multi-step algebraic or arithmetic complexity  such as layered percentage changes, fractional relationships, or proportional reasoning  while preserving the original problem format, clarity, and conciseness. Ensure all quantities, units, and percentage bases are explicitly stated, specifying whether percentages apply to original or updated amounts after prior changes. Clearly and unambiguously specify the exact sequence of all changes (e.g., first X\% are removed, then Y\% wilt, then Z are added) to avoid temporal or logical ambiguity. Design numerical values to produce answers that require careful calculation, including non-terminating decimals or close approximations, so that small computational errors yield plausible but incorrect results, thereby increasing cognitive challenge. Limit the length increase to no more than 50\% relative to the original problem, verifying this quantitatively by comparing word or character counts and pruning redundant or non-essential information to maintain conciseness. Balance added complexity with accessibility by avoiding excessive layering that could overwhelm the target student level, and embed subtle linguistic cues  such as transitional phrases (after these changes,subsequently)  to guide reasoning without explicit hints or meta-text. Before finalizing, internally simulate a student reasoning path to confirm the problem remains fair, solvable, and yields a single clear numerical answer without unnecessary rounding ambiguity. Justify each added element by confirming it introduces a necessary algebraic or arithmetic step that deepens understanding without causing confusion or undue cognitive load. Avoid introducing advanced or unfamiliar concepts outside the original domain and curriculum scope, and ensure all terminology and phrasing are unambiguous and age-appropriate. Finally, exclude any meta-text or instructional language from the output, producing a clean, standalone problem statement that reflects these enhancements
\end{promptboxNL}

\begin{promptboxNL}{3}
\ttfamily\small
Increase the mathematical difficulty of this problem by thoughtfully introducing multi-step reasoning, layered relationships, and fractional arithmetic involving dynamic quantities that change as the problem progresses (e.g., fractions of remaining amounts). Ensure all new elements such as rounding, intermediate calculations, subtractions, or fractional shares are clearly defined with precise sequencing and operational clarity to avoid ambiguity or unfair traps. For example, specify exactly when and how rounding occurs relative to other steps (e.g., immediately after the percentage increase, round down to the nearest whole number before any subtraction). Use unambiguous, standard phrasing for fractional relationships (e.g., Andy paid half the \$3 cost of the fries rather than paid half the fries) and explicitly clarify whether costs or quantities are per unit or totals. Balance the increase in difficulty so it challenges students appropriately for the target level without causing undue confusion or frustration; avoid complexity that is too subtle to impact reasoning or too overwhelming to be fair. Maintain consistent terminology, units, and currency symbols throughout. Do not introduce concepts outside the intended domain (e.g., avoid advanced probability, calculus, or abstract algebra); ensure all new elements fit within standard arithmetic and algebraic reasoning suitable for the audience. Include explicit instructions or clarifications within the problem text to guide students and reduce ambiguity, especially around multi-step or layered relationships. Anticipate common student errors by incorporating complexity that encourages critical thinking but avoids unfair or misleading traps. Limit the length increase to no more than 50\% of the original problem length, keeping the problem concise and focused. Ensure the final problem ends with exactly one clear, unambiguous numerical question that directly relates to the introduced complexity and is solvable with careful reasoning. Return ONLY the revised problem text without any introductory phrases, meta-comments, evaluations, rubrics, or extraneous commentary. Before finalizing, internally verify that the problem instructions are unambiguous, the sequence of operations is clear, and the problem remains fair and solvable.
\end{promptboxNL}

\begin{promptboxNL}{4}
\ttfamily\small
Modify this problem by adding a single, cohesive, and logically integrated condition that meaningfully increases its mathematical difficulty without altering the original format. The added condition must be fully contextualized within the problem scenario, clearly explaining any changes in quantities, timing, or relationships such as why and how chairs are added or removed, from which groups or tables, and how these changes relate to the original static counts. Avoid vague or ambiguous phrases (e.g., immediately after, make up the rest) and instead use precise, natural language that aligns with typical student understanding, explicitly defining all temporal terms, units, and dependencies.Introduce difficulty through multi-step algebraic reasoning, conditional logic, or combining continuous rates with discrete events in a way that requires students to set up and solve equations or inequalities. Ensure the added condition targets common student error patterns (e.g., confusion between discrete and continuous changes, unit mismatches) and aligns with standard algebraic concepts such as linear equations or systems of equations, avoiding advanced topics beyond the intended curriculum.Temporal or dynamic changes should be intrinsically linked to the problem context and mathematically meaningful; avoid introducing unexplained or extraneous temporal dynamics that do not naturally fit the original domain or require assumptions beyond the student scope. Favor static or conditionally defined quantities over ambiguous temporal rates, and if periodic events are included, specify explicit timing and dependencies clearly.The revised problem must remain concise, with total length not exceeding 1.5 times the original word count. Internally verify this length constraint by counting words or tokens, and if exceeded, simplify or condense language while preserving clarity and difficulty. Prioritize readability by breaking complex information into clear, stepwise sentences without parenthetical asides, and maintain consistent terminology and units throughout.End the problem with exactly one clear numerical question that explicitly requires engagement with the added condition; the question must be impossible to answer correctly without applying the new complexity. Avoid superficial or disconnected additions.Internally generate a rationale explaining how the added condition increases difficulty, addressing clarity, fairness, curriculum alignment, and how it targets common student misunderstandings. This rationale should also confirm adherence to length constraints and the natural integration of the new condition within the original problem context.Do not include any meta-text, examples, or rationale in the output problem statement itself; these are for internal use only to guide generation and support downstream evaluation and refinement.
\end{promptboxNL}

\section{Qualitative Results}
\label{app:qualititative-results}

\subsection{Examples of generated problems}
\label{app:generated-problem-variations}

\subsubsection{Problems generated with Opt-NL}
\label{sec:appendix:generated_problem_opt_nl}

\subsection*{Original Problem}
Liza bought 10 kilograms of butter to make cookies. She used one-half of it for chocolate chip cookies, one-fifth of it for peanut butter cookies, and one-third of the remaining butter for sugar cookies. How many kilograms of butter are left after making those three kinds of cookies?

\subsection*{Generated Questions}

\begin{enumerate}
    \item Liza bought 10 kilograms of butter to make cookies. She used one-half of it for chocolate chip cookies, then used one-fifth of the remaining butter for peanut butter cookies, and then used one-third of the butter remaining after those two kinds for sugar cookies. How many kilograms of butter are left after making those three kinds of cookies?
    
    \item Liza bought 10 kilograms of butter to make cookies. She used one-half of it for chocolate chip cookies, then she used one-fifth of the original 10 kilograms for peanut butter cookies. After those two uses, she used one-third of the remaining butter for sugar cookies, and then she gave away an amount equal to one-fourth of the butter she had used for chocolate chip cookies. How many kilograms of butter are left after these actions?
    
    \item Liza bought 10 kilograms of butter to make cookies. She used one-half of it for chocolate chip cookies. Then she used one-fifth of the remaining butter for peanut butter cookies, and after that she used one-third of the remaining butter for sugar cookies. Finally, 0.10 kilograms of butter were accidentally spilled and lost while transferring. How many kilograms of butter are left after all of those events?
    
    \item Liza bought 10.0 kilograms of butter to make cookies. She set aside 0.2 kilograms for sampling, then used one-half of the remaining butter for chocolate chip cookies, one-fifth of the butter remaining after that for peanut butter cookies, and one-third of the butter remaining after those two batches for sugar cookies. How many kilograms of butter are left after making those three kinds of cookies?
    \item Liza bought 10 kilograms of butter to make cookies. First she used one-half of the 10 kg for chocolate chip cookies. Next she used one-fifth of the remaining butter (after making the chocolate chip cookies) for peanut butter cookies. Finally, after those two uses she used one-third of the butter then remaining for sugar cookies. How many kilograms of butter are left after making those three kinds of cookies?
    
    \item Liza bought 10 kilograms of butter to make cookies. She used one-half of it for chocolate chip cookies, then used one-fifth of the remaining butter for peanut butter cookies, and finally used one-third of the remaining butter after that for sugar cookies. How many kilograms of butter are left after making those three kinds of cookies?

    \item Liza bought 10 kilograms of butter to make cookies. She used one-half of it for chocolate chip cookies. Then she used one-fifth of the remaining butter for peanut butter cookies. From what was left after that, she used one-third for sugar cookies. Finally, she set aside an amount equal to one-tenth of the remaining butter (after the sugar cookies) to give to a neighbor. How many kilograms of butter are left after all these uses?

    \item Liza bought 10 kilograms of butter to make cookies. She used one-half of it for chocolate chip cookies, then used one-fifth of the remaining butter (after making the chocolate chip cookies) for peanut butter cookies, and then used one-third of the remaining butter (after making the peanut butter cookies) for sugar cookies. How many kilograms of butter are left after making those three kinds of cookies?

    \item Liza bought 10 kilograms of butter to make cookies. She first used one-half of it for chocolate chip cookies. Then she used one-fifth of the remaining butter for peanut butter cookies. Next she used one-third of the new remaining butter for sugar cookies. Finally, she added an extra amount to the sugar cookies equal to 20\% of the butter she had used for the peanut butter cookies, taking that extra from whatever butter was left at that point. How many kilograms of butter are left after all of these uses?

    \item Liza bought 10 kilograms of butter to make cookies. She used one-half of the original 10 kg for chocolate chip cookies, then used one-fifth of the remaining butter for peanut butter cookies, and then used one-third of the butter remaining after those two uses for sugar cookies. How many kilograms of butter are left after making those three kinds of cookies?

\end{enumerate}

\subsubsection{Problems generated with \ourmethod}
\label{sec:appendix:generated_problem_our_method}

\subsection*{Seed Question}
Liza bought 10 kilograms of butter to make cookies. She used one-half of it for chocolate chip cookies, one-fifth of it for peanut butter cookies, and one-third of the remaining butter for sugar cookies. How many kilograms of butter are left after making those three kinds of cookies?

\subsection*{Generated Questions}

\begin{enumerate}
    \item Marco had 10 assorted candies. He gave one-half of them to friends, one-fifth to neighbors, and then gave one-third of the remaining candies to his sister. How many candies remain?
    
    \item Ethan bought 10 kilograms of flour for baking. He used one-half of it for bread, one-fifth of it for muffins, and then used one-third of the remaining flour for pancakes. How many kilograms of flour remain after these uses?
    
    \item Marco had 10 candies to share. He gave half of them away as chocolate candies, one-fifth as peanut candies, and then one-third of the remaining candies as sugar candies. How many candies does he have left?
    
    \item Ethan purchased 10 liters of white paint to repaint his garage. He used one-half of it on the garage door, one-fifth of it on the trim, and then one-third of the remaining paint for touch-ups. How many liters of paint are left after those three uses?
    
    \item Marcus had 10 candies. He set aside one-half of them as chocolate candies and one-fifth of them as peanut candies. From what was left, he used one-third for sugar candies. How many candies remain after making those three kinds?
    
    \item Maya had 10 candies to share. She gave one-half of them as chocolate candies, one-fifth as peanut candies, and then gave one-third of the remaining candies as strawberry candies. How many candies remain after those three distributions?
    
    \item Sam bought some candies. He gave one-half of them as chocolate candies, one-fifth as peanut candies, and then gave one-third of the remaining candies as sugar candies. After those distributions, 2 candies were left. How many candies did Sam buy at the start?
    
    \item Marco had 10 bags of candy. He gave one-half of them away to friends, one-fifth of them to classmates, and then donated one-third of what was left to a charity. How many bags of candy remain?
    
    \item Marco had 10 pounds of trail mix. He gave one-half of it to a group of hikers, one-fifth of it to the camp kitchen, and then packed one-third of the remaining trail mix into small snack bags. How many pounds of trail mix are left?
    
    \item Marco bought 10 candy bars to share. He gave one-half of them as chocolate treats, one-fifth as peanut-flavored treats, and then gave one-third of the remaining bars as sugar-flavored treats. How many candy bars are left after those distributions?
\end{enumerate}

\subsection{Comparison of problems generated via different representations.}\label{app:representation-comparison}

\subsubsection*{Example 1}

\begin{tcolorbox}[colback=white,colframe=black!50!white,title=Original,after skip=2pt,before skip=2pt]
Natalia sold clips to 48 of her friends in April, and then she sold half as many clips in May. How many clips did Natalia sell altogether in April and May?
\end{tcolorbox}
\begin{tcolorbox}[colback=coralheading!20!white,colframe=smtcolor,title=SMT Mode,after skip=2pt,before skip=2pt]
In April, a bookstore sold 49 copies of a popular novel. In May, the store sold fewer copies, specifically 48\% of the number sold in April, but since partial copies can't be sold, the number of copies sold in May is rounded down to the nearest whole number. The total number of copies sold over these two months is the sum of the copies sold in April and May. How many copies were sold in total during April and May?
\begin{lstlisting}
(declare-fun april_sold () Int)
(declare-fun may_sold () Int)
(declare-fun total_sold () Int)
; April sales fixed at 49 to introduce a prime number and slightly increase complexity
(assert (= april_sold 49))
; May sales set to be (to_int (* 0.48 april_sold)) to simulate a 48% discount, using to_int truncation
(assert (= may_sold (to_int (* 0.48 april_sold))))
; total_sold is sum of april_sold and may_sold with nested addition emphasizing arithmetic order
(assert (= total_sold (+ april_sold (+ may_sold 0))))
(check-sat)
(get-value (total_sold))
\end{lstlisting}
\end{tcolorbox}

\begin{tcolorbox}[colback=sympycolor!20!white,colframe=sympycolor,title=Sympy Mode,after skip=2pt,before skip=2pt]
Liam printed 48 posters for a local event, and each poster cost him \$1 to make. He didn’t owe any tax and didn’t qualify for any bulk discount, but later he decided to print extra posters equal to half of what he had already made. How much money did Liam spend in total on all the posters?
\begin{lstlisting}
# base quantity (e.g., number of items)
num_items = sp.Integer(48)
# unit price (rational for exact arithmetic)
price_per_item = sp.Rational(1, 1)
# subtotal: sum of item costs
subtotal = num_items * price_per_item  # 48 * 1 = 48
# tax on subtotal (set to zero here, expressed as a rational rate)
tax_rate = sp.Rational(0, 1)
tax_amount = subtotal * tax_rate  # tax_amount = 0
# discount computed via integer division of quantities (floor)
bulk_count_for_discount = sp.floor(num_items / sp.Integer(100))  # yields 0
discount_amount = sp.Rational(0, 1) * bulk_count_for_discount  # zero discount
# explicit half amount corresponding to 'm = a/2'
half_of_subtotal = sp.Rational(1, 2) * subtotal  # 24
# final total: subtotal + tax + extra half - discount
final_total = subtotal + tax_amount + half_of_subtotal - discount_amount
# simplified numeric answer assigned to 'answer'
answer = sp.simplify(final_total)
\end{lstlisting}
\end{tcolorbox}

\begin{tcolorbox}[colback=nlcolor!20!white,colframe=nlcolor,title=NL Mode,after skip=2pt,before skip=2pt]
Natalia sold clips to 48 different friends in April; each friend bought either 1 clip or 2 clips. In May she sold half as many clips as she did in April. The number of friends who bought 2 clips in April equals two-thirds of the number of clips she sold in May. How many clips did Natalia sell altogether in April and May?
\end{tcolorbox}

\subsubsection*{Example 2}

\begin{tcolorbox}[colback=white,colframe=black!50!white,title=Original,after skip=2pt,before skip=2pt]
Weng earns \$12 an hour for babysitting. Yesterday, she just did 50 minutes of babysitting. How much did she earn?
\end{tcolorbox}

\begin{tcolorbox}[colback=coralheading!20!white,colframe=smtcolor,title=SMT Mode,after skip=2pt,before skip=2pt]
Alex wants to buy a wallet that costs \$120. Alex has already saved some money, which is slightly less than half the cost of the wallet. Specifically, Alex's savings amount to 48\% of the wallet's price, rounded down to the nearest whole dollar. Alex's parent contributes \$18 towards the purchase. Additionally, Alex's grandparent contributes an amount that is about 2.33 times the parent's contribution, calculated by multiplying the parent's contribution by 7 and then dividing by 3, using only whole dollars. After adding Alex's savings, the parent's contribution, and the grandparent's contribution, there is still some money needed to buy the wallet. How much more money does Alex need to fully pay for the wallet?

\begin{lstlisting}
(declare-fun wallet_cost () Int)
(declare-fun initial_savings () Int)
(declare-fun parent_contribution () Int)
(declare-fun grandparent_contribution () Int)
(declare-fun total_contributions () Int)
(declare-fun remaining_need () Int)

(assert (= wallet_cost 120)) ;; increased cost for difficulty
(assert (= initial_savings (to_int (* 0.48 wallet_cost)))) ;; slightly less than half, truncated
(assert (= parent_contribution 18)) ;; increased parent contribution
(assert (= grandparent_contribution (div (* parent_contribution 7) 3))) ;; grandparent contributes about 2.33 times parent, integer division
(assert (= total_contributions (+ parent_contribution grandparent_contribution)))
(assert (= remaining_need (- wallet_cost (+ initial_savings total_contributions))))
(check-sat)
(get-value (remaining_need))
\end{lstlisting}

\end{tcolorbox}

\begin{tcolorbox}[colback=sympycolor!20!white,colframe=sympycolor,title=Sympy Mode,after skip=2pt,before skip=2pt]
A contractor charges \$12 per hour and worked for 50 minutes on a job. Because the time worked differs from a half hour, the contractor applies a performance adjustment factor equal to 1 plus three times the square of the difference (in hours) between the time worked and a half hour. In addition, the contractor adds a small fixed extra amount of time equal to the sum of the squares of 1 and 2 divided by 120 hours. The billable time is the adjusted time worked plus that small extra amount, and the total payment is the hourly rate multiplied by the billable time. What is the total payment in dollars for this job?

\begin{lstlisting}
# hourly rate in dollars per hour
hourly_rate = sp.Rational(12)
# minutes worked
minutes_worked = sp.Rational(50)
# convert minutes to hours
hours_worked = minutes_worked / sp.Rational(60)
# deviation from a half hour used to create a small quadratic adjustment factor
deviation_from_half = hours_worked - sp.Rational(1, 2)
# nonlinear adjustment factor = 1 + 3*(deviation)^2 (pure rational arithmetic)
adjustment_factor = sp.Rational(1) + sp.Rational(3) * (deviation_from_half ** 2)
# closed-form sum of squares for n=2: n(n+1)(2n+1)/6
n_terms = sp.Integer(2)
sum_of_squares = n_terms * (n_terms + 1) * (2 * n_terms + 1) / sp.Integer(6)
# small additive quadratic term based on the closed-form sum scaled rationally
additive_quadratic_term = sum_of_squares / sp.Rational(120)
# effective billable hours combining nonlinear factor and quadratic summation term
effective_hours = hours_worked * adjustment_factor + additive_quadratic_term
# total earnings = rate * effective_hours
earnings = hourly_rate * effective_hours
# final numeric rational answer
answer = sp.simplify(earnings)
\end{lstlisting}
\end{tcolorbox}

\begin{tcolorbox}[colback=nlcolor!20!white,colframe=nlcolor,title=NL Mode,after skip=2pt,before skip=2pt]
Sarah earns \$12 per hour babysitting. One day, she babysat for 60 minutes. Since there are 60 minutes in one hour, Sarah calculates how many hours she babysat by converting the minutes into hours. Then, she multiplies the number of hours she babysat by her hourly wage to find out how much money she earned that day. How much money did Sarah earn from babysitting?
\end{tcolorbox}


\end{document}